%% file: manuscript.tex
\documentclass[journal]{IEEEtran}
\usepackage{amsmath,amsfonts,amssymb}
\usepackage{algorithmic}
\usepackage{algorithm}
\usepackage{array}
\usepackage[caption=false,font=footnotesize,labelfont=rm,textfont=rm]{subfig}
\usepackage{textcomp}
\usepackage{stfloats}
\usepackage{verbatim}
\usepackage{graphicx}
\usepackage{cite}

\usepackage{mathrsfs}
\usepackage{bbm}
\usepackage{booktabs}
\usepackage{balance}
\usepackage{multirow}
\usepackage{footnote}
\usepackage{color}
\usepackage{xcolor}
\usepackage{diagbox}
\usepackage{enumitem}

\usepackage{url}
\usepackage{bm}
\usepackage{threeparttable}
\usepackage{xr-hyper}
\usepackage[colorlinks,linkcolor=blue,urlcolor=blue,citecolor=blue]{hyperref}

\begin{document}

\title{In-Place Panoptic Radiance Field Segmentation with Perceptual Prior for 3D Scene Understanding}

\author{Shenghao~Li
\thanks{Shenghao Li email: lch94102@gmail.com}
}

\markboth{Journal of \LaTeX\ Class Files,~Vol.~14, No.~8, August~2021}%
{Shell \MakeLowercase{\textit{et al.}}: A Sample Article Using IEEEtran.cls for IEEE Journals}

\IEEEpubid{0000--0000/00\$00.00~\copyright~2021 IEEE}

\maketitle

\begin{abstract}

Accurate 3D scene representation and panoptic understanding are essential for applications such as virtual reality, robotics, and autonomous driving. However, challenges persist with existing methods, including precise 2D-to-3D mapping, handling complex scene characteristics like boundary ambiguity and varying scales, and mitigating noise in panoptic pseudo-labels. This paper introduces a novel perceptual-prior-guided 3D scene representation and panoptic understanding method, which reformulates panoptic understanding within neural radiance fields as a linear assignment problem involving 2D semantics and instance recognition. Perceptual information from pre-trained 2D panoptic segmentation models is incorporated as prior guidance, thereby synchronizing the learning processes of appearance, geometry, and panoptic understanding within neural radiance fields. An implicit scene representation and understanding model is developed to enhance generalization across indoor and outdoor scenes by extending the scale-encoded cascaded grids within a reparameterized domain distillation framework. This model effectively manages complex scene attributes and generates 3D-consistent scene representations and panoptic understanding outcomes for various scenes. Experiments and ablation studies under challenging conditions, including synthetic and real-world scenes, demonstrate the proposed method's effectiveness in enhancing 3D scene representation and panoptic segmentation accuracy.

\end{abstract}


\section{Introduction} \label{section-introduction}

Building upon the foundation of 3D scene reconstruction and representation, the flexible and reliable panoptic understanding of 3D scenes is deemed essential for the application of 3D scene reconstruction in fields such as virtual reality, robot navigation, and autonomous driving. The joint learning of geometric appearance representation and semantic instance segmentation within the scene is crucial for enhancing the flexibility of online scene reconstruction and representation. In the realm of 2D panoptic segmentation, images are segmented into background categories (Stuff) and foreground categories (Thing). This field has experienced rapid development through the evolution of deep learning models and the release of large-scale annotated 2D datasets, leading to the emergence of advanced 2D panoptic segmentation models \cite{cheng2022masked, li2023mask, yu2022k}, which demonstrate high generalization in understanding panoptic images observed in real-world scenarios.

The panoptic understanding of a scene is fundamentally based on 3D panoptic segmentation. From a 2D perspective, panoptic segmentation integrates two computer vision tasks: semantic segmentation and instance segmentation \cite{kirillov2019panoptic}. For a single observed image, semantic segmentation assigns a category label to each pixel, while instance segmentation detects and segments each target instance. Panoptic segmentation unifies these approaches by assigning category labels to every pixel and detecting and segmenting each instance within the observed image. Subsequently, 3D panoptic segmentation extends this unified process into three-dimensional space.
\IEEEpubidadjcol

\begin{figure}[t]
	\centering
	\includegraphics[width=\columnwidth]{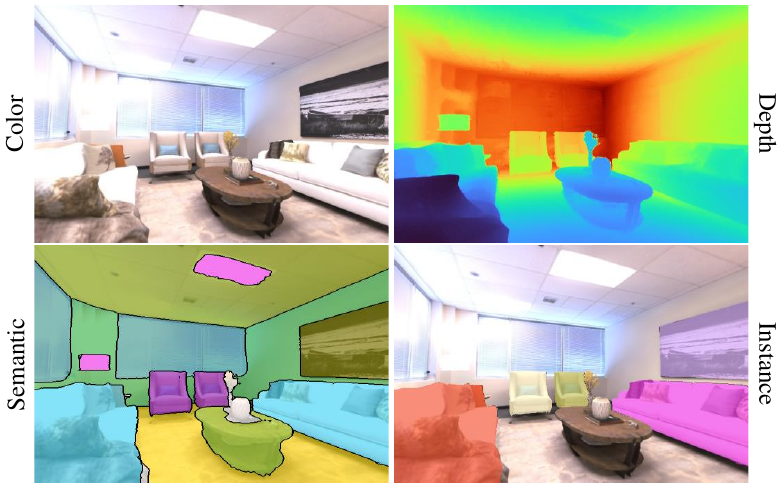}
	\caption{Illustration of 3D Scene Representation and Panoptic Understanding}
	\label{fig1-vis-intro}
\end{figure}

However, despite the strong performance of 2D panoptic segmentation methods on individual images, they frequently encounter viewpoint-dependent segmentation errors and lack consistent classification across different views. This inconsistency renders them unsuitable for tasks requiring 3D continuity and consistency from multiple viewpoints. Consequently, deriving scene panoptic segmentation results with 3D continuity and consistency based on 2D segmentation outputs remains a significant research challenge. Specifically, existing methods that attempt to bridge 2D segmentation and 3D scene representation for panoptic understanding encounter the following primary issues:

\textbf{3D Mapping Accuracy}: The construction of accurate 2D-to-3D mapping is fundamental for 3D scene representation and panoptic understanding. This necessitates the integration of observed 2D image information and its panoptic segmentation with visual sensor pose estimation methods to develop 3D reconstruction and representation models, as well as panoptic segmentation models of the target scene.

\textbf{Characteristics of Scenes}: Addressing various features of target scenes, such as boundary ambiguity and varying scales, requires the design of a highly generalizable scene parameterization system. Establishing efficient implicit scene representation and panoptic understanding models is essential for improving the accuracy and robustness of 3D scene representation and panoptic understanding.

\textbf{Pseudo-Label Noise}: In the process of learning 2D-to-3D panoptic understanding, pseudo-labels for semantic and instance information are generated by performing panoptic segmentation on observed 2D images. The quality of these pseudo-labels directly impacts the accuracy of scene representation and panoptic understanding. Given that the 2D panoptic segmentation results may inherently contain errors and noise, effectively mitigating the noise in panoptic pseudo-labels is critical for obtaining accurate 3D scene representation and panoptic understanding models.

Recent advancements in 3D scene understanding utilizing Neural Radiance Fields (NeRF) have facilitated 3D panoptic comprehension of scenes~\cite{fu2022panoptic, kundu2022panoptic, wang2022dm, zhi2021place, siddiqui2023panoptic}. Some of these works~\cite{fu2022panoptic, wang2022dm} rely on 2D or 3D ground truth segmentation labels, which are challenging to acquire for real-world scene reconstruction and representation tasks. The PNF method~\cite{kundu2022panoptic} achieves 2D-to-3D panoptic understanding by leveraging pre-trained 2D semantic segmentation models and 3D bounding box detection methods; however, its performance is limited by the generalization capability of the pre-trained 3D detection models~\cite{siddiqui2023panoptic}. The state-of-the-art Panoptic-Lifting approach~\cite{siddiqui2023panoptic} employs pre-trained 2D panoptic segmentation methods for 3D mapping within the NeRF framework and addresses pseudo-label noise during this process. Nevertheless, this method primarily focuses on indoor scenes with well-defined boundaries and does not account for boundary ambiguity in outdoor environments.

To overcome these challenges, a perceptual prior guided 3D scene representation and panoptic understanding method is proposed in this paper. The proposed method formulates the panoptic understanding of neural radiance fields as a linear assignment problem from 2D pseudo labels to 3D space. By incorporating high-level features from pre-trained 2D panoptic segmentation models as prior guidance, the learning processes of appearance, geometry, semantics, and instance information within the neural radiance field are synchronized. Furthermore, the implicit scene representation model is extended and updated by constructing a novel implicit scene representation and understanding model using scale-encoded cascaded grids within a reparameterization domain distillation framework. Consequently, the proposed method generates 3D-consistent scene representations and panoptic understanding for both indoor and outdoor environments, as illustrated in Fig.~\ref{fig1-vis-intro}.

The main contributions of the proposed method are as follows:
\begin{enumerate}
\item A novel multi-task learning framework for panoptic segmentation in neural radiance fields is introduced, wherein the geometry, appearance, semantics, and instance-level information of every point in the 3D scene are represented and modeled based on appearance observation data and 2D panoptic labels of the scene;
\item A new implicit scene representation and understanding model is developed, capable of adapting to target scenes with complex characteristics such as boundary ambiguity, thereby providing consistent 3D reconstruction and panoptic understanding across diverse target scenes;
\item Perceptual information from pre-trained 2D panoptic segmentation models is incorporated as guidance. By utilizing a patch-based ray sampling strategy, the joint optimization of geometry, appearance, semantics, and instance information within the neural radiance field is achieved, enhancing the accuracy of representation and understanding;
\item Extensive experiments are conducted, including comparisons with state-of-the-art algorithms on multiple datasets and qualitative visual evaluations. The main theoretical components and modules involved are also evaluated through ablation studies.
\end{enumerate}

\section{Related Work} \label{section-related-work}

Scene reconstruction and representation methods can geometrically reconstruct and faithfully restore indoor and outdoor scenes \cite{barron2021mipnerf,barron2022mipnerf360}. However, with the development of mobile robots and artificial intelligence, their application carriers are shifting towards Intelligent Agents (IA). When planning tasks in their environment, intelligent agents need not only to perceive the geometry of the scene but also to make corresponding judgments in combination with the semantic and instance information in the scene, responding to the needs of human users. Therefore, scene reconstruction and representation methods should not only focus on the representation of geometry and appearance but also perceive semantics and instances, thereby empowering the development of intelligent agents and human-computer interaction in environment-oriented applications~\cite{siddiqui2023panoptic}.

\begin{figure}[t] 
    \centering
    \includegraphics[width=\columnwidth]{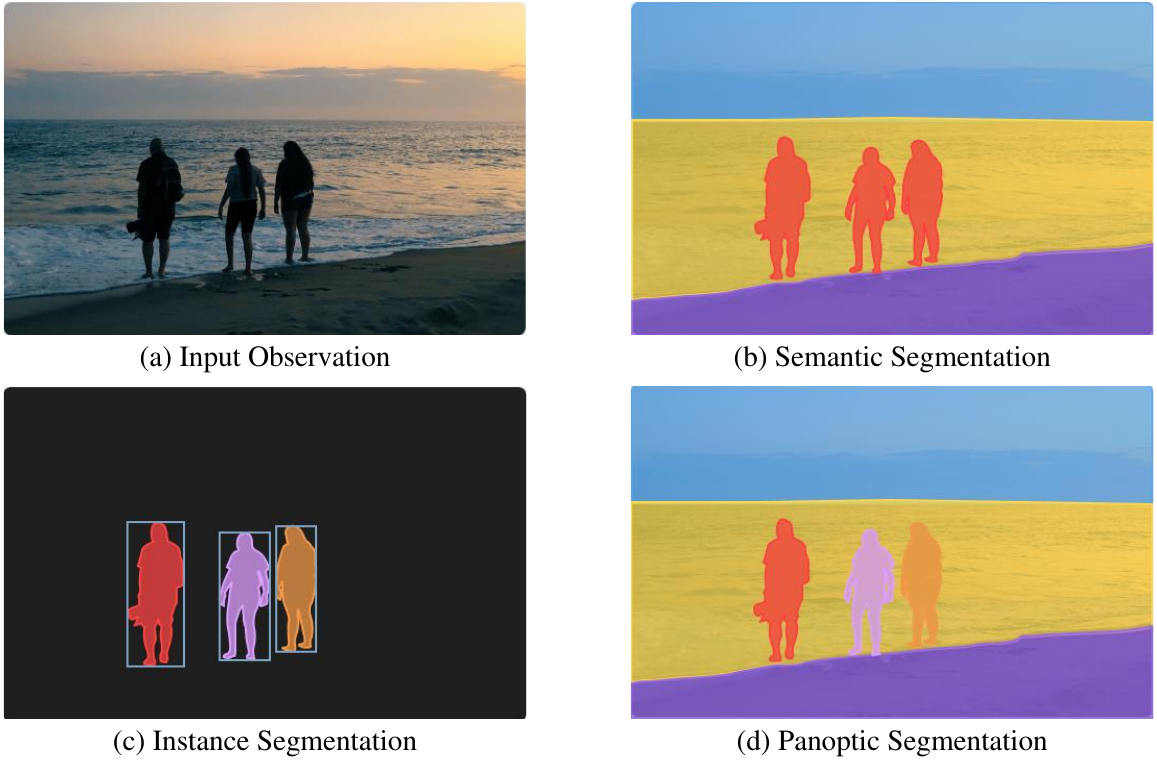} 
    \caption{The Categories of Segmentation in Scene Understanding} 
    \label{fig2-panoptic-segmentation} 
\end{figure}

In the research and application of computer vision and robotics, segmentation tasks can generally be divided into semantic segmentation, instance segmentation, and panoptic segmentation based on the target object and processing process, as shown in Fig.~\ref{fig2-panoptic-segmentation}. Among them, semantic segmentation mainly assigns a unique category to each pixel in the observation and does not distinguish between instances within the category. As shown in Fig.~\ref{fig2-panoptic-segmentation}(a), semantic segmentation separates the beach, sky, ocean, and humans but does not distinguish between the three humans. Instance segmentation detects specific targets in the image and extracts their masks. As shown in Fig.~\ref{fig2-panoptic-segmentation}(b), instance segmentation detects and distinguishes three humans. Panoptic segmentation combines the characteristics of semantic segmentation and instance segmentation, classifying the things in the scene and distinguishing the same things \cite{kirillov2019panoptic}, as shown in Fig.~\ref{fig2-panoptic-segmentation}(d).

Panoptic segmentation covers the main processes of semantic segmentation and instance segmentation, and semantic segmentation can divide the total categories into foreground categories (Things) and background categories (Stuff) \cite{caesar2018coco}. The main differences between the two are: 
\begin{enumerate} 
    \item \textbf{Shape}: Foreground objects often have unique shapes, such as vehicles, animals, and mobile phones, while background objects are shapeless, such as sky, water, and grassland. 
    \item \textbf{Size}: The size of foreground objects often changes little, while the size of background objects varies greatly. 
    \item \textbf{Parts}: Foreground objects can generally be decomposed into recognizable component parts, such as animal limbs, while background objects cannot be, for example, a part of water is still water. 
    \item \textbf{Instances}: Foreground objects are independent and countable instances, while background objects are uncountable and cannot be clearly divided into instances. 
    \item \textbf{Texture}: Background objects often have more texture characteristics, such as lawns and sky, while foreground objects do not. 
\end{enumerate}

It is worth noting that there are still some special cases in panoptic segmentation that can be considered foreground or background objects under different conditions. For example, humans are generally considered foreground objects, but when a large crowd appears and is not the focus, the crowd can also be considered a background object.

Many 3D scene understanding methods achieve scene segmentation and understanding by attaching 3D labels to the 3D geometric representation reconstructed from the scene. Hermans et al. attached semantic labels to the dense indoor point cloud reconstructed from RGB-D observations, forming a 3D semantic map \cite{hermans2014dense}. McCormac et al. proposed SemanticFusion \cite{mccormac2017semanticfusion}, which uses a convolutional neural network to extract Surfel semantic information based on ElasticFusion and maps and fuses semantic prediction results from multiple different viewpoints to obtain a semantic map. Runz et al. further proposed MaskFusion \cite{runz2018maskfusion}, which can recognize, segment, and assign semantic category labels to targets in the scene during the simultaneous localization and mapping process of RGB-D SLAM. McCormac et al. proposed Fusion++ \cite{mccormac2018fusion++}, which initializes a Truncated Signed Distance Function by introducing MaskRCNN \cite{he2017mask} instance segmentation during the learning process and jointly updates the reconstruction of geometric structures and the representation of semantic understanding during the scene reconstruction process. Narita et al. divided scene understanding into foreground objects and background objects and proposed PanopticFusion \cite{narita2019panopticfusion}, which performs individual segmentation of any target in the foreground while densely predicting category labels in the background area.

Han et al. proposed a geometry occupancy-aware 3D instance segmentation method, OccuSeg \cite{han2020occuseg}, which uses a multi-task learning method to regress instance segmentation and geometric structure from the feature dimension, achieving advanced results on multiple indoor datasets. Qi et al. proposed PointNet++ \cite{qi2017pointnet++} for the segmentation problem of 3D point clouds, which learns local features of 3D point clouds based on metric space distance and recursively divides the input point cloud set. Huang et al. proposed TextureNet \cite{huang2019texturenet}, which introduces a consistent four-direction rotation symmetric field for the segmentation problem of textured meshes, thereby extracting directed graph blocks from mesh textures based on sampling points and performing 3D semantic segmentation. Schult et al. proposed DualConvMesh-Net \cite{schult2020dualconvmesh}, which considers both the geodesic information and geometric information of the 3D scene. Subsequently, Hu et al. proposed VMNet \cite{hu2021vmnet}, which voxelizes the input 3D mesh surface, extracts geodesic domain and Euclidean domain information for the grids and voxels respectively, and fuses them based on the attention mechanism, improving the effect of 3D semantic segmentation. Hu et al. proposed a Bidirectional Projection Network (BPNet) \cite{hu2021bidirectional}, which jointly understands 2D and 3D spaces in an end-to-end structure, enabling information exchange between 2D UNet \cite{ronneberger2015unet} and 3D MinkowskiUNet \cite{choy20194d} and enhancing scene understanding effects. Nekrasov et al. proposed Mix3D \cite{nekrasov2021mix3d}, a data augmentation method for 3D segmentation algorithms, which mixes the input 3D scene representation and fuses global visual text information relationships and local features to solve local ambiguity problems and global information overfitting problems in scene segmentation. Rozenberszki et al. proposed using the CLIP \cite{radford2021learning} feature encoding pre-training model for 3D semantic segmentation, using text encoding for pre-training the 3D point feature encoder, and training the decoder based on 3D ground truth labels. Li et al. proposed Panoptic-phnet \cite{li2022panoptic}, which performs panoptic segmentation for geometric reconstruction based on lidar, locates instance targets by learning cluster pseudo score maps, and realizes information interaction between foreground target points based on KNN-Transformer, thereby realizing information sharing across multiple perception domains and fusion with voxel features. Robert et al. proposed a multi-view feature aggregation method based on an end-to-end implicit model for large-scale 3D scene segmentation problems, fusing 2D and 3D information for large-scale scene semantic segmentation \cite{robert2022learning}. These methods use 3D ground truth labels in the learning process and have a profound impact on derivative applications of 3D scene understanding, including 3D object classification \cite{wu20153d}, 3D object detection and localization \cite{chen2020scanrefer,niemeyer2021giraffe, sun2020scalability}, 3D semantic and instance segmentation \cite{armeni2017joint,behley2019semantickitti,liao2023kitti}, 3D feasibility prediction \cite{deng20213d,li2019putting,wang2021synthesizing}, and so on.

Another approach to learning 3D scene understanding is based on 2D supervised segmentation signals. Genova et al. proposed using 2D image supervision information combined with multi-view geometry to fuse into 3D semantic pseudo-labels, thereby training a 3D semantic segmentation model \cite{genova2021learning}. Kundu et al. targeted the 3D grid reconstructed from RGB-D observation data, rendered multiple virtual viewpoint 2D images, and trained a 2D semantic segmentation model, then aggregated the 2D segmentation results and applied them back to the 3D surface, thereby achieving semantic understanding of the 3D scene \cite{kundu2020virtual}. Sautier et al. proposed using a self-supervised pre-training method for 3D perception of autonomous driving multimodal data (images and lidar) \cite{sautier2022image}, by training a 3D network through 2D pixel features with a self-supervised distillation method. Wang et al. proposed Detr3d \cite{wang2022detr3d}, which extracts 2D features from multi-view observation images, then uses a set of sparse 3D query objects to index these 2D features, uses visual sensor transformation matrices to link 3D coordinates to multi-view images, and finally predicts the bounding box of the query object, introducing a set loss to measure the loss value compared with the ground truth. Liu et al. trained a 3D segmentation network through contrastive learning by utilizing the pairing relationship between 3D points and 2D pixels \cite{liu2021contrastive}. These methods are called closed-set methods because they limit the number of categories of scientific semantic labels or example labels. The representative advanced work among them, Panoptic-Lifting \cite{siddiqui2023panoptic}, proposed learning the 3D panoptic radiance field for indoor scene 3D panoptic understanding using a high-performance pre-trained 2D panoptic segmentation model and dealing with the segmentation noise carried by the pre-trained segmentation model from the dimension of probability.

On the other hand, with the rapid development of recent large-scale visual-language models \cite{jia2021scaling, radford2021learning, alayrac2022flamingo}, multimodal models have made important progress in zero-shot scene understanding tasks based on 2D images, greatly improving the robustness of scene understanding, including recognizing long-tail objects in images. Many works establish open-vocabulary models through the cross-correlation between dense image features and large language model encoding, thereby achieving image segmentation based on arbitrary text labels \cite{gu2021open, ghiasi2022scaling, he2023open, ma2022open, zhou2022extract}. Ghiasi et al. proposed OpenSeg \cite{ghiasi2022scaling}, addressing the problem that open vocabulary cannot achieve pixel-level segmentation of visual content by proposing a visual grouping method, grouping pixels before learning the correspondence between visual features and semantic information, thereby achieving visual semantic segmentation learning with text captions as supervision signals. L{\"u}ddecke et al. proposed CLIPSeg \cite{luddecke2022image}, which is based on the encoding extracted by the text and visual Transformer in CLIP \cite{radford2021learning}, specifically training a decoder to segment query images. Ha et al. further proposed Semantic Abstraction \cite{ha2022semantic}, achieving open-vocabulary small-scale scene understanding and completion based on single-evidence RGB-D observation data, and requiring ground truth data as supervision signals during retraining. Peng et al. proposed OpenScene, which does not rely on annotated data of 3D point clouds but distills knowledge from the pre-trained visual-language large model CLIP \cite{radford2021learning}, combined with 3D fusion of 2D information to achieve open-vocabulary understanding of 3D scenes \cite{Peng2023OpenScene}.

In general, the mainstream work in 3D scene understanding focuses on semantic segmentation of the target scene, learning semantic labels on the basis of 3D scene representation models to achieve 3D understanding of the scene. In this process, high-performance scene representation methods can effectively improve the quality of scene understanding, and compared to the higher-cost 3D semantic labels, mapping the 2D semantic labels obtained from 2D image segmentation to 3D semantic pseudo-labels can greatly improve development efficiency. In addition, by introducing additional prior information, such as the features of pre-trained 2D segmentation models or the conceptual knowledge of large language models, the accuracy and diversity of scene understanding can be significantly improved. Moreover, scene understanding can be further expanded on the basis of semantic information to distinguish between different instances, thereby achieving panoptic understanding of the scene.

\section{Methodology} \label{section-methodology}

The proposed perceptual-prior-guided 3D scene representation and panoptic understanding method aims to achieve panoptic segmentation results with 3D consistency from arbitrary viewpoints within the scene. This objective is accomplished by utilizing observation images along with the intrinsic and extrinsic parameters of the target scene's visual sensor through joint learning with an implicit scene representation and understanding model. In addition to synthesizing color maps and depth maps from free viewpoints, semantic probability distribution maps $\mathbf{U} \in \mathbb{R}^{H \times W \times U}$ for semantic segmentation and instance probability distribution maps $\mathbf{V} \in \mathbb{R}^{H \times W \times V}$ for instance segmentation are rendered, where $H$ and $W$ denote the image's height and width, and $U$ and $V$ represent the number of categories for semantic and instance segmentation, respectively. The semantic segmentation categories encompass both foreground objects ($U_T$) and background objects ($U_S$).

\begin{figure*}[t] 
    \centering 
    \includegraphics[width=1.9\columnwidth]{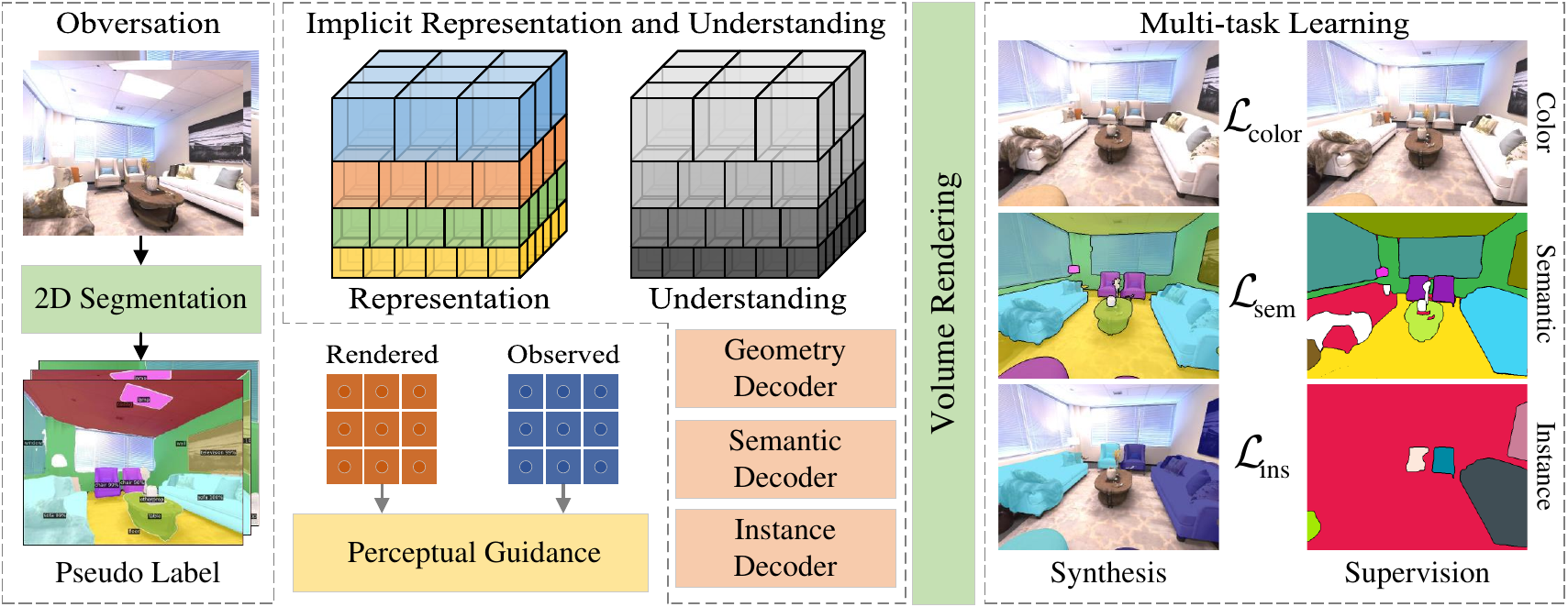} 
    \caption{Overview of the Perceptual Prior Guided 3D Scene Representation and Panoptic Understanding. By employing an understanding feature grid and dual decoders, we jointly learn geometric, appearance, semantic, and instance information. Perceptual prior guidance further enhances the alignment between geometric and semantic features, enabling accurate and consistent 3D panoptic segmentation from arbitrary viewpoints.}
    \label{fig3-panoptic-nerf} 
\end{figure*}

As illustrated in Fig.~\ref{fig3-panoptic-nerf}, panoptic segmentation maps corresponding to the observed RGB data are first extracted using Mask2Former\cite{cheng2022masked}, a pre-trained 2D panoptic segmentation network. These maps include semantic pseudo-labels and instance pseudo-labels, which serve as supervision signals for the subsequent learning of scene representation and panoptic understanding. Subsequently, the scene semantic and instance segmentation modules are introduced in the implicit scene representation model. This model comprises an understanding feature grid and two decoders, thereby forming a novel implicit scene representation and understanding framework for the scene's panoptic radiance field, denoted as $\mathcal{S}: (\mathbf{x}, \mathbf{d}) \mapsto (\sigma, \mathbf{c}, \mathbf{u}, \mathbf{v})$. In this notation, $\mathbf{x} \in \mathbb{R}^3$ and $\mathbf{d} \in \mathbb{R}^3$ represent the coordinates and shooting direction of 3D points in the target scene, $\sigma \in \mathbb{R}$ denotes the volumetric density, $\mathbf{c} \in \mathbb{R}^3$ the directional color, $\mathbf{u} \in \mathbb{R}^U$ the semantic category vector, and $\mathbf{v} \in \mathbb{R}^V$ the instance vector of each 3D point. The implicit scene representation and understanding model then performs multi-task joint learning under the guidance of various supervision signals. Through volume rendering, it synthesizes viewpoints to obtain color maps, depth maps, semantic segmentation maps, and instance segmentation maps from any free viewpoint within the target scene, thereby achieving comprehensive 3D representation and understanding. Additionally, during the representation learning process, perceptual-guided regularization is employed to enhance the association between appearance geometry and semantic instances in the feature space. This is achieved by introducing perceptual guidance from the pre-trained panoptic segmentation network, thereby improving the multi-task learning capabilities of the implicit scene representation and understanding model. The visual sensor observation poses of the RGB observation data can be calculated using the online process described in our previous work \cite{li2024representing} or through offline methods such as COLMAP\cite{schonberger2016colmap}.

\subsection{Observation Preprocessing} \label{subsection-observation-preprocess}

The perceptual-prior-guided 3D scene representation and panoptic understanding method is initiated with a sequence of RGB images that have been calibrated with the intrinsic and extrinsic parameters of the visual sensor. Each input observation image is denoted as $\hat{\mathbf{C}} \in \mathbb{R}^{H \times W \times 3}$, where $H$ and $W$ correspond to the height and width of the observation frame, respectively. The intrinsic and extrinsic matrices of the visual sensor are predicted using the approaches described in our previous works~\cite{li2023s2ld} and \cite{li2024representing}, utilizing checkerboard calibration. Upon inputting the observation image $\hat{\mathbf{C}}$ into a pre-trained 2D panoptic segmentation network, the corresponding semantic supervision map $\hat{\mathbf{U}} \in \mathbb{R}^{H \times W \times U}$ and instance supervision map $\hat{\mathbf{V}} \in \mathbb{R}^{H \times W \times V}$ are obtained. The vectors in these semantic and instance supervision maps are employed as pseudo-labels for supervision throughout the learning process of 3D scene representation and panoptic understanding.

\begin{figure}[t!] 
    \centering 
    \includegraphics[width=1\columnwidth]{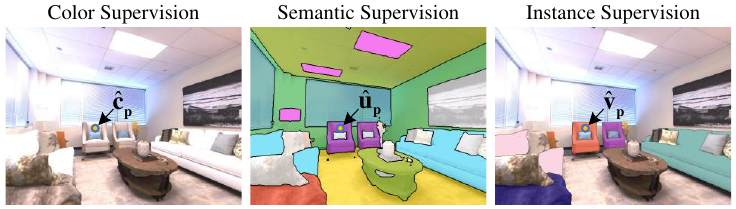} 
    \caption{Illustration of Observation Data Preprocessing. This process involves calibrating the intrinsic and extrinsic parameters of the visual sensor and generating semantic and instance supervision maps using a pre-trained panoptic segmentation network.}
    \label{fig4-preprocess} 
\end{figure}

3D points within the scene are positioned through interval sampling along rays that are projected from the origin $\mathbf{o} \in \mathbb{R}^3$ of the visual sensor in the direction $\mathbf{d}$. These points are denoted as $\mathbf{p}(t) = \mathbf{o} + t\mathbf{d}$, where $t \in \mathbb{R}$ represents the distance value used for sampling along the ray. Each pixel corresponding to a ray in the observation image is associated with an RGB color pseudo-label vector $\hat{\mathbf{c}}_\mathbf{p} \in \mathbb{R}^3$, a semantic category pseudo-label vector $\hat{\mathbf{u}}_\mathbf{p} \in \mathbb{R}^U$, and an instance category pseudo-label vector $\hat{\mathbf{v}}_\mathbf{p} \in \mathbb{R}^V$. It should be noted that $\hat{\mathbf{v}}_\mathbf{p}$ is defined for the $U_T$ foreground object classes. The perceptual-prior-guided 3D scene representation and panoptic understanding method employs Mask2Former\cite{cheng2022masked}, a pre-trained 2D panoptic segmentation network, to generate the semantic category pseudo-label vector $\hat{\mathbf{u}}_\mathbf{p}$ and the instance category pseudo-label vector $\hat{\mathbf{v}}_\mathbf{p}$ based on the provided observation data, as illustrated in Fig.~\ref{fig4-preprocess}. The elements within the semantic and instance category pseudo-label vectors represent the probabilities corresponding to each category for the respective point.

\subsection{Implicit Representation and Understanding Model} \label{subsection-implicit-representation}

For accurate 3D scene representation and panoptic understanding, a high-performance implicit representation model is crucial for predicting the volume density, color, semantic category, and instance category of points in 3D space based on their positions and directions. The perceptual-prior-guided 3D scene representation and panoptic understanding method extends the implicit scene representation model employed in \cite{li2024representing}. The updated and expanded Panoptic Radiance Field implicit scene representation and understanding model is denoted as $\mathcal{S}: (\mathbf{x}, \mathbf{d}) \mapsto (\sigma, \mathbf{c}, \mathbf{u}, \mathbf{v})$, where $\mathbf{x} \in \mathbb{R}^3$ and $\mathbf{d} \in \mathbb{R}^3$ represent the coordinates and shooting direction of a 3D point within the target scene, respectively. The volume density and directional color of the 3D point are represented by $\sigma \in \mathbb{R}$ and $\mathbf{c} \in \mathbb{R}^3$, respectively, while $\mathbf{u} \in \mathbb{R}^U$ and $\mathbf{v} \in \mathbb{R}^V$ denote the semantic and instance probability distribution vectors of the 3D point, respectively.

\begin{figure*}[t] 
    \centering 
    \includegraphics[width=1.9\columnwidth]{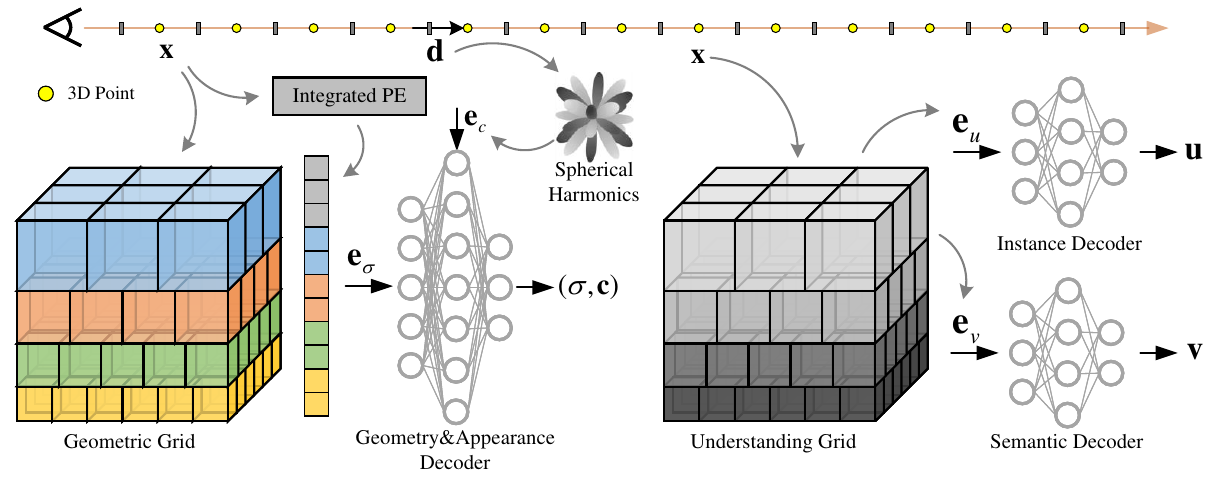} 
    \caption{Overview of the implicit scene representation and understanding model $\mathcal{S}$. Geometric and understanding feature grids are integrated, and multi-layer perceptrons are employed to generate semantic and instance probability distributions. The geometry, appearance, semantic, and instance decoders are jointly trained to predict volume density, directional color, semantic and instance probability distributions.}
    \label{fig5-panoptic-radiance-field} 
\end{figure*}

The overall schematic of the perceptual-prior-guided 3D scene representation and panoptic understanding method is depicted in Fig.~\ref{fig5-panoptic-radiance-field}. A multi-resolution voxel grid~\cite{li2024representing} is retained as a geometric feature grid, and position encoding along with spherical harmonic functions are jointly integrated to compute the volume density encoding $\mathbf{e}_\sigma$ and color encoding $\mathbf{e}_c$ for each 3D point $\mathbf{x}$ in the direction $\mathbf{d}$. The volume density $\sigma$ and color $\mathbf{c}$ of the scene 3D point are predicted through geometric and appearance decoders, respectively. For the prediction of semantic and instance probability distributions, an understanding feature grid, which is smaller in scale than the geometric feature grid, is constructed to provide encoding for both the semantic decoder and instance decoder. This understanding encoding primarily captures the positional information of the 3D scene point, ensuring consistency between the semantic encoding $\mathbf{e}_u$ and instance encoding $\mathbf{e}_v$ as they share the same feature grid. The semantic decoder and instance decoder, composed mainly of multi-layer perceptrons (MLPs), take the semantic and instance encodings as inputs and output the semantic category probability distribution vector $\mathbf{u} \in \mathbb{R}^{U}$ and instance probability distribution vector $\mathbf{v} \in \mathbb{R}^V$ for each scene 3D point $\mathbf{x}$. Here, $\left\{u_i\right\} \in \mathbf{u}$ and $\left\{v_i\right\} \in \mathbf{v}$ represent the probabilities of their corresponding semantic and instance categories, respectively.

\subsection{Multi-Task Learning}\label{subsection-multi-task-learning}

For a given sampling cone ray $\mathbf{p}(t)=\mathbf{o} + t\mathbf{d}$ in the world coordinate system, the $i$-th 3D spatial point $\mathbf{x}_i$ sampled along the ray is considered. This point is predicted by the proposed method to possess a volumetric density of $\sigma_i$, a color vector of $\mathbf{c}_i$, a semantic probability distribution of $\mathbf{u}_i$, an instance probability distribution of $\mathbf{v}_i$, and a sampling endpoint distance of $t_i$. The color value, semantic category, and instance category of the corresponding pixel on the imaging plane can be computed as:
\begin{equation} 
\begin{aligned}
    \mathcal{C}(\mathbf{p}) = \sum_{i=1}^{N} w_i \mathbf{c}_{i}, \ \ 
    \mathcal{U}(\mathbf{p}) &= \text{softmax}\left(\sum_{i=1}^{N} w_i \mathbf{u}_{i}\right), \\
    \mathcal{V}(\mathbf{p}) &= \text{softmax}\left(\sum_{i=1}^{N} w_i \mathbf{v}_i \right), \label{eq-panoptic-rendering} 
\end{aligned}
\end{equation} 
where $\left\{t_i\right\}$ and $\left\{w_{i}\right\}, i = 1, 2, \cdots, N$ represent the distance values and corresponding weights sampled along the cone projection direction, respectively, and $N$ denotes the number of samples. The weights $w_{i}$ are defined by the following formula:
\begin{equation} 
    w_i = \exp \left(-\sum_{j=1}^{i-1} \sigma_{j} \Delta t_{j}\right)\left(1-\exp \left(-\sigma_{i} \Delta t_{i}\right)\right), 
\end{equation} 
where $\Delta t_{i} = t_{i+1} - t_{i}$ signifies the sampling bucket length. In Equation \eqref{eq-panoptic-rendering}, $\mathcal{C}(\cdot)$ denotes the color rendering function, while $\mathcal{U}(\cdot)$ and $\mathcal{V}(\cdot)$ represent the volume rendering functions for the semantic probability distribution and instance probability distribution, respectively. The semantic category of the pixel corresponding to this ray is determined by $u^* = \arg \max (\mathcal{U}(\mathbf{p}))$. If $u^*$ corresponds to a foreground object, the unique instance category of this pixel is obtained by calculating $v^* = \arg \max (\mathcal{V}(\mathbf{p}))$. Consequently, the 3D-consistent panoptic segmentation result of this pixel is represented as $(u^*, v^*)$.

The color loss function in the panoptic radiance field is defined using the Charbonnier Loss between the rendered color map $\mathbf{C} \in \mathbb{R}^{H \times W \times 3}$ and the observed color map $\hat{\mathbf{C}} \in \mathbb{R}^{H \times W \times 3}$. This loss function is expressed by the following equation:
\begin{equation} 
    \mathcal{L}_{\mathrm{color}}(\mathbf{C}, \hat{\mathbf{C}}) = \operatorname{mean}\left(\sqrt{(\mathbf{C} - \hat{\mathbf{C}})^2 + \epsilon^2}\right), 
\end{equation} 
where $\operatorname{mean}(\cdot)$ denotes the averaging operation, and $\epsilon = 10^{-4}$ serves as the boundary constant for the Charbonnier loss function.

The scene semantic and instance segmentation modules, as illustrated in Fig.~\ref{fig3-panoptic-nerf}, are primarily constructed using multi-layer perceptrons (MLPs). These modules receive the coordinates of 3D scene points as inputs and output semantic probability distribution vectors $\mathbf{u}$ and instance probability distribution vectors $\mathbf{v}$, respectively. For a set of sampled rays $\mathbf{P}$, the loss function for 3D semantic segmentation is defined as the weighted cross-entropy loss applied to the semantic probability distributions:
\begin{equation} 
    \mathcal{L}_\text{sem}(\mathbf{P}) = -\frac{1}{|\mathbf{P}|} \sum_{\mathbf{p} \in \mathbf{P}} \lambda_\mathbf{p} \sum_{i=1}^{U} \hat{u}_i \log(u_i), 
\end{equation} 
where $\lambda_\mathbf{p}$ denotes the confidence weight of the pseudo-label associated with the pixel's semantic segmentation corresponding to ray $\mathbf{p}$, $u_i$ represents the probability of the $i$-th semantic category in the vector $\mathbf{u}$, $\hat{u}_i$ is the $i$-th element of the one-hot encoded 2D semantic segmentation pseudo-label vector, and $|\mathbf{P}|$ indicates the total number of sampled rays.

Similarly, the loss function for 3D instance segmentation is defined as the weighted cross-entropy loss applied to the instance probability distribution:
\begin{equation} 
    \mathcal{L}_{\text{ins}}(\mathbf{P}) = -\frac{1}{|\mathbf{P}|} \sum_{\mathbf{p} \in \mathbf{P}} \lambda_\mathbf{p} \sum_{i=1}^{V} \hat{v}_i \log(v_i), 
\end{equation} 
where $v_i$ represents the probability of the $i$-th instance category in the vector $\mathbf{v}$, and $\hat{v}_i$ is the $i$-th element of the one-hot encoded 2D instance segmentation pseudo-label vector.

To mitigate the noise present in the pseudo-labels of the panoptic segmentation results from the observed data, the proposed method introduces a segmentation consistency loss function. By clustering, a set of rays sharing the same pseudo-label semantic category is sampled, denoted as $\mathbf{P}'$, and the segmentation consistency loss function is defined as follows:

\begin{equation} 
    \mathcal{L}_{\mathrm{seg}}(\mathbf{P}') = -\frac{1}{|\mathbf{P}'|} \sum_{\mathbf{p} \in \mathbf{P}'} \lambda_\mathbf{p} \sum_{i=1}^U \tilde{u}_i \log (u_i),
\end{equation} 

where $\tilde{u}_i$ is the $i$-th element in $\tilde{\mathbf{u}}$, and $\tilde{\mathbf{u}}$ represents the one-hot probability distribution corresponding to the dominant pseudo-label category within the sampled ray group. Since all rays in $\mathbf{P}'$ share the same semantic category, $\mathcal{L}_{\mathrm{seg}}$ encourages the implicit scene representation and understanding model to predict consistent semantic probability distributions for 3D space points with identical semantic labels, thereby enhancing the 3D consistency of scene understanding.

\section{Perceptual Prior Guidance} \label{subsection-perceptual-prior-guidance}

\begin{figure}[t] 
    \centering 
    \includegraphics[width=\columnwidth]{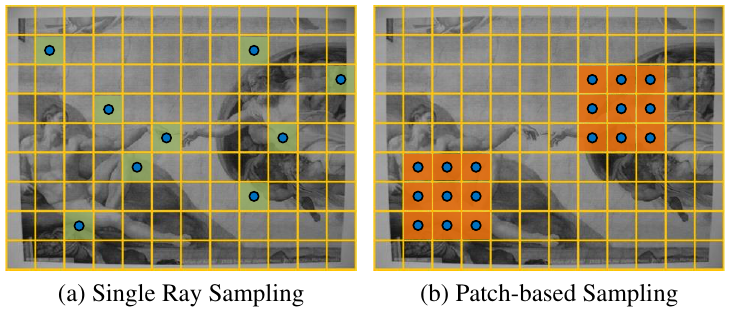} 
    \caption{Illustration of Patch-Based Sampling. Patch-based ray sampling aggregates neighboring rays into patches, enabling joint sampling that preserves inter-ray correlations.} 
    \label{fig6-patch-sample} 
\end{figure}

The proposed method learns a 3D scene understanding model by integrating 2D panoptic segmentation results with an implicit scene representation and understanding model through multi-task learning. However, during the scene representation learning process, challenges such as noise in 2D pseudo-labels and inconsistencies among appearance, geometry, semantics, and instance predictions emerge. To address these issues, perceptual guidance and various regularization terms derived from patch-based ray sampling are applied, enhancing the robustness and consistency of the representation learning and resulting in a more accurate 3D scene representation and panoptic understanding model.

Patch-based ray sampling, illustrated in Fig.~\ref{fig6-patch-sample}, is compared to single ray sampling. In the single ray sampling approach shown in Fig.~\ref{fig6-patch-sample}(a), each pixel of the observed image is treated as an independent ray for sampling, thereby neglecting inter-ray correlations. In contrast, the patch-based ray sampling method depicted in Fig.~\ref{fig6-patch-sample}(b) involves joint sampling of rays within patches, significantly preserving the correlations between rays and facilitating subsequent perceptual guidance and regularization based on multi-ray patches.

The perceptual guidance is integrated by incorporating a pre-trained feature extraction network, which directs the 3D scene representation using high-dimensional feature space information. A group of rays sampled through patch-based sampling is denoted as $\mathbf{P}$, and the perceptual guidance is defined by the following equation:
\begin{equation} 
    \mathcal{L}_{\mathrm{feat}}(\mathbf{P})= \frac{1}{|\mathbf{P}|}\sum_{\mathbf{p} \in \mathbf{P}} \left\| \mathcal{F}_\mathrm{percep}\left(\mathbf{C}_\mathbf{P}\right) - \mathcal{F}_\mathrm{percep}\left(\hat{\mathbf{C}}_\mathbf{P}\right) \right\|_2, 
\end{equation} 
where $\mathbf{C}_\mathbf{P}$ and $\hat{\mathbf{C}}_\mathbf{P}$ represent the rendered and observed color patches of the patch-sampled rays, respectively, $\|\cdot\|_2$ denotes the $\ell_2$ norm, and $\mathcal{F}_\mathrm{percep}(\cdot)$ signifies the prior feature extraction function. To enhance consistency between appearance geometry and panoptic understanding within the panoptic radiance field, the feature guidance prior utilizes the Swin-L architecture from Mask2Former~\cite{cheng2022masked} to establish the prior feature extraction function $\mathcal{F}_\mathrm{percep}(\cdot)$.

Building upon patch-based ray sampling, two regularization terms are introduced in the multi-task learning framework of scene representation and panoptic understanding, further guiding the learning process of the 3D scene representation and panoptic understanding model. These include the regularization loss functions $\mathcal{L}_\mathrm{tv}$ and $\mathcal{L}_\mathrm{disp}$. The purpose of $\mathcal{L}_\mathrm{tv}$ is to promote similarity between feature vectors in adjacent nodes within the feature grid, thereby reconstructing a more compact geometric structure. This regularization term is defined by the following formula: 
\begin{equation} 
    \mathcal{L}_{\mathrm{tv}}\left(\mathcal{G}\right)=\operatorname{mean}\left(\sum^{\mathcal{G}} \left\|\boldsymbol{\xi}_i - \boldsymbol{\xi}_{i+1}\right\|_2\right), 
\end{equation} 
where $\operatorname{mean}(\cdot)$ denotes the averaging operation, $\mathcal{G}$ is the multi-resolution voxel grid of the appearance geometry feature grid and semantic segmentation feature grid described in Section \ref{subsection-implicit-representation}, and $\boldsymbol{\xi}_i$ and $\boldsymbol{\xi}_{i+1}$ are the feature vectors carried by adjacent nodes in the voxel grid $\mathcal{G}$.

The term $\mathcal{L}_\mathrm{disp}$ regularizes the geometric structure of the 3D scene representation based on disparity by constraining the disparity term (the reciprocal of the depth value) within a small range to mitigate the ghosting phenomenon in viewpoint rendering. This regularization term is defined by the following formula: 
\begin{equation} 
    \mathcal{L}_{\mathrm{disp}}(\mathbf{P}) = \frac{1}{|\mathbf{P}|}\sum_{\mathbf{p} \in \mathbf{P}}\sum_{i=1}^{N} w_i \frac{1}{t_i}, 
\end{equation} 
where $\left\{t_{i}\right\}$ and $\left\{w_{i}\right\}, i = 1, 2, \cdots, N$ represent the distance values sampled along the projection direction of the rays in $\mathbf{P}$ and the corresponding weights, respectively, $N$ is the number of samples, and $\Delta t_{i} = t_{i+1} - t_{i}$ is the bucket length, consistent with Eq.~\eqref{eq-panoptic-rendering}.

The overall loss function of the 3D scene representation and panoptic understanding model is defined by the following formula: 
\begin{equation} 
\begin{aligned}
    \mathcal{L}_\mathrm{total} = &\mathcal{L}_\mathrm{color} + \alpha_\mathrm{distill}\sum\mathcal{L}_\mathrm{distill} + \alpha_\mathrm{sem}\mathcal{L}_\mathrm{sem} + \alpha_\mathrm{ins}\mathcal{L}_\mathrm{ins} + \\
    &\alpha_\mathrm{seg}\mathcal{L}_\mathrm{seg} + \alpha_\mathrm{feat}\mathcal{L}_\mathrm{feat} + \alpha_\mathrm{reg}\left(\mathcal{L}_\mathrm{tv} + \mathcal{L}_\mathrm{disp}\right), 
\end{aligned}
\end{equation} 
where $\sum\mathcal{L}_\mathrm{distill}$ denotes the distillation loss function~\cite{li2024representing} of the multi-level structure in the implicit scene representation and understanding model. The balance hyperparameters $\alpha_\mathrm{distill}$, $\alpha_\mathrm{sem}$, $\alpha_\mathrm{ins}$, $\alpha_\mathrm{seg}$, $\alpha_\mathrm{feat}$, and $\alpha_\mathrm{reg}$ correspond to each loss function, and their specific values are determined through experimental tuning, as detailed in Section \ref{subsubsection-setting-details}.


\section{Experimental Results and Analysis}
The proposed method is experimentally evaluated across multiple datasets, and a series of ablation studies are conducted to validate the comprehensive performance of the 3D scene representation and panoptic understanding, as well as the effectiveness of each individual module.

\subsection{Experimental Setting}

\subsubsection{Dataset Description}

A diverse set of datasets is utilized to evaluate the proposed method, ensuring comprehensive experimentation. These datasets include Replica~\cite{straub2019replica}, HyperSim~\cite{roberts2021hypersim}, ScanNet~\cite{dai2017scannet}, and KITTI-360~\cite{liao2023kitti}. Replica, HyperSim, and ScanNet pertain to indoor scenes, whereas KITTI-360 is designated for outdoor scenes. The object categories employed in panoptic segmentation experiments are detailed in Table~\ref{table-thing-stuff}, encompassing 21 object types for indoor settings as described in \cite{cheng2022masked} and 20 object types for outdoor settings as outlined in \cite{liao2023kitti}.

\subsubsection{Evaluation Metrics}

The proposed method is primarily assessed using the following evaluation metrics, where an upward arrow ($\uparrow$) signifies that higher values denote better performance, and vice versa:

\textbf{Peak Signal-to-Noise Ratio (PSNR$\uparrow$)}: This metric quantifies the quality of the reconstructed luminance by measuring the difference between the rendered color image and the ground truth image.

\textbf{Mean Intersection over Union (mIOU$\uparrow$)}: This metric evaluates the accuracy of semantic segmentation by calculating the intersection over union between the rendered semantic map and the ground truth semantic map.

\textbf{Scene-level Panoptic Quality (PQ$^\text{scene}$$\uparrow$)}: This metric assesses the quality of panoptic segmentation within the target scene by comparing the degree of alignment between the rendered semantic and instance maps with the supervised semantic and instance maps.

\textbf{Scene-level Segmentation Quality (SQ$^\text{scene}$$\uparrow$)}: This metric measures the segmentation accuracy of panoptic segmentation in the target scene by evaluating the differences in segmentation between the rendered semantic and instance maps and the supervised semantic and instance maps.

\textbf{Scene-level Retrieval Quality (RQ$^\text{scene}$$\uparrow$)}: This metric determines the retrieval effectiveness of panoptic segmentation in the target scene by comparing the retrieval discrepancies between the rendered semantic and instance maps and the supervised semantic and instance maps.

\subsubsection{Baseline Methods}

Experimental comparisons are conducted with advanced 3D semantic segmentation methodologies, including SemanticNeRF~\cite{zhi2021place}, as well as 3D panoptic segmentation approaches such as DM-NeRF~\cite{wang2022dm}, PNF~\cite{kundu2022panoptic}, and Panoptic-Lifting~\cite{siddiqui2023panoptic}. For evaluations involving the outdoor autonomous driving dataset KITTI-360, additional viewpoint synthesis algorithms augmented by 2D panoptic segmentation, specifically NeRF~\cite{mildenhall2021nerf} and MipNeRF~\cite{barron2021mipnerf}, are also employed.

\subsubsection{Experimental Details} \label{subsubsection-setting-details}

The proposed method is assessed on a desktop system equipped with an Intel i7-10700K CPU and an RTX3090 GPU. The appearance geometry component of the implicit scene representation and understanding model is maintained consistently with Ours.L as outlined in~\cite{li2024representing}. A voxel grid with single-dimensional features at a resolution of 16 is utilized to construct the semantic instance feature grid. Both semantic and instance decoders are realized using 256-dimensional MLPs. The balance hyperparameters for the loss function are established as $\alpha_\mathrm{distill}=1.2$, $\alpha_\mathrm{sem}=0.1$, $\alpha_\mathrm{ins}=0.1$, $\alpha_\mathrm{seg}=0.12$, $\alpha_\mathrm{feat}=0.2$, and $\alpha_\mathrm{reg}=0.001$, following parameter tuning experiments. In the comparative analyses for 3D scene panoptic understanding, the number of semantic categories is assigned as 22 for indoor scenes ($U=22$) and 21 for outdoor scenes ($U=21$), inclusive of the blank category (Void). The maximum number of instances is set to 65 for both indoor and outdoor scenes ($V=65$).

\subsection{3D Representation and Understanding Benchmark}

The proposed method is evaluated against the baseline methods across multiple indoor and outdoor scene datasets in the 3D representation and understanding benchmark. A comprehensive experimental evaluation is conducted, assessing aspects such as appearance, geometry, semantics, and instances of scene representation.

\subsubsection{Benchmark on the Replica Dataset} 

\input{tables/table_02_replica.tex}

The Replica dataset is utilized to compare the proposed method with the baseline methods in indoor scenes. As the Replica dataset is synthetic, ideal shooting conditions are maintained, and the scenes consist of smaller room scales with clear boundaries. The benchmark is performed at a resolution of $512 \times 512$, aligning with the experimental criteria outlined in \cite{siddiqui2023panoptic}.

As shown in Table~\ref{table-benchmark-replica}, the proposed method achieves superior results in appearance geometry representation (PSNR), 3D semantic segmentation (mIOU), and 3D panoptic segmentation (PQ$^\text{scene}$) compared to the baseline methods. The enhancement in appearance geometry representation primarily originates from the implicit scene representation and understanding model $\mathcal{S}$. Additionally, the improvements in 3D semantic and instance segmentation are attributed to the increased 3D segmentation consistency facilitated by perceptual-prior-guided regularization and the segmentation consistency loss function. However, the proposed method exhibits slightly lower performance than Panoptic-Lifting and PNF with ground truth bounding boxes (PNF+GT.Box) in terms of the SQ$^\text{scene}$ metric. This reduction is due to the more substantial regularization imposed on the semantic understanding module under conditions where boundaries are well-defined and shooting conditions are optimal. Nevertheless, the proposed method continues to demonstrate robust scene representation and understanding capabilities. In subsequent experiments, the proposed method will be compared with the baseline methods in more challenging target scenes.

\subsubsection{Benchmark on the HyperSim Dataset} 

\input{tables/table_01_hypersim.tex}

The benchmark on the HyperSim dataset is performed in indoor scenes. As a synthetic dataset, HyperSim provides ideal shooting conditions, featuring larger room scales with clear scene boundaries. These experiments are conducted at a resolution of $512 \times 512$, aligning with the experimental criteria outlined in \cite{siddiqui2023panoptic}.

As presented in Table~\ref{table-benchmark-hypersim}, the proposed method (Ours) achieves high metrics in appearance representation (PSNR) and panoptic segmentation (PQ$^\text{scene}$, SQ$^\text{scene}$, RQ$^\text{scene}$) compared to the baseline methods. In terms of semantic segmentation (mIOU), the proposed method ranks second, trailing only the advanced algorithm Panoptic-Lifting. The elevated PSNR scores indicate that the proposed method facilitates more accurate reconstruction and representation of target scenes, thereby enhancing the learning of 3D semantic and instance segmentation. Additionally, the high mIOU and PQ$^\text{scene}$ metrics demonstrate that the proposed method effectively learns a 3D panoptic segmentation implicit model with substantial accuracy and 3D consistency in larger-scale indoor environments.

\subsubsection{Benchmark on the ScanNet Dataset} 

\input{tables/table_03_scannet.tex}

The benchmark on the ScanNet dataset is performed in more challenging indoor scenes. Being a real-world indoor dataset, ScanNet introduces observed data with inherent noise and includes apartment scenes of varying scales. These experiments are performed at a resolution of $256 \times 256$, with panoptic segmentation resampled to $512 \times 512$, consistent with the experimental criteria described in \cite{siddiqui2023panoptic}.

As illustrated in Table~\ref{table-benchmark-scannet}, the proposed method surpasses baseline methods in appearance geometry representation and semantic segmentation, as evidenced by higher PSNR and mIOU scores. Furthermore, the panoptic segmentation metrics PQ$^\text{scene}$ and RQ$^\text{scene}$ indicate performance close to state-of-the-art baseline methods. These results demonstrate that the proposed method effectively handles 3D reconstruction and panoptic understanding tasks in apartment indoor scenes across different scales.

\subsubsection{Benchmark on the KITTI-360 Dataset} 

\input{tables/table_05_kitti360_cus.tex}

The benchmark on the KITTI-360 dataset is performed in challenging outdoor scenes. As a real-world outdoor driving dataset, KITTI-360 encompasses large-scale outdoor environments without defined boundaries. Five sequences from the KITTI-360 dataset are selected for evaluation, conducted at the original resolution of $1408 \times 376$. In these experiments, baseline methods MipNeRF \cite{barron2021mipnerf} and NeRF \cite{mildenhall2021nerf} utilize reproduced code incorporating 3D panoptic segmentation, while Mask2Former \cite{cheng2022masked} and Panoptic-Lifting \cite{siddiqui2023panoptic} employ their official open-source implementations.

As shown in Table~\ref{table-benchmark-kitti360-cus}, the proposed method delivers strong performance in both 3D reconstruction representation and panoptic understanding within outdoor scenes. The implicit scene representation and understanding model demonstrates significant improvements in appearance geometry representation, attributed to the method’s design considerations for boundary ambiguity in outdoor environments (as detailed in Section \ref{subsection-implicit-representation}), as reflected by the PSNR metric. Building upon this robust appearance geometry representation, the proposed method achieves superior semantic segmentation and panoptic segmentation quality, as indicated by the mIOU and PQ$^\text{scene}$ metrics, through the combined effects of feature-prior-guided regularization and the segmentation consistency loss function.

\subsubsection{Analysis of Running Efficiency} 
\input{tables/table_07_efficiency.tex}

Table~\ref{table-runtime-analysis} presents the running efficiency analysis of the proposed method and the baseline methods. The analysis primarily reflects the rendering efficiency by measuring the time required to process the same number (2048) of rays using an RTX3090 GPU. It is evident from the table that the proposed method achieves viewpoint synthesis of the target scene's appearance geometry and semantic instances with low rendering time.

\begin{figure}[h!]
    \centering
    \includegraphics[width=1\columnwidth]{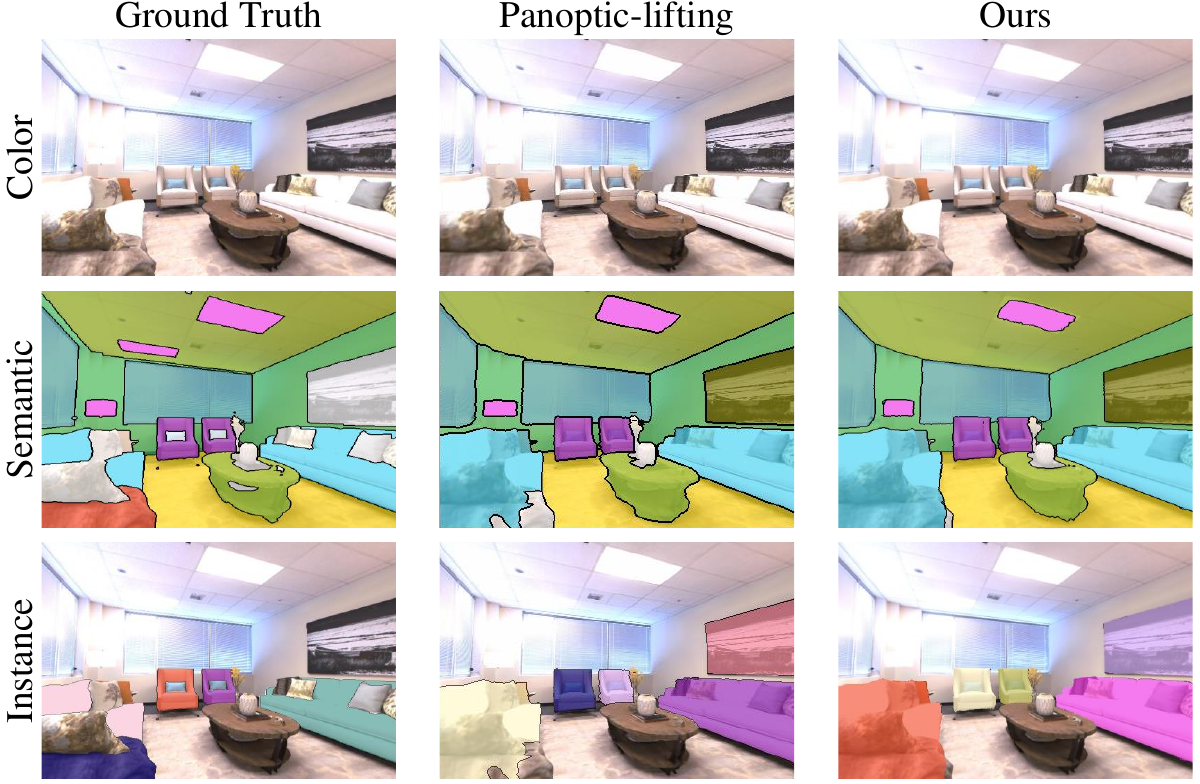}
    \caption{Illustration of Qualitative Results on the Replica Dataset. The proposed method effectively enhances indoor scene understanding by utilizing feature-prior-guided regularization and segmentation consistency loss, resulting in accurate and 3D-consistent panoptic segmentation.}
    \label{fig7-vis-replica}
\end{figure}

\subsubsection{Visualization of Qualitative Experimental Results}

\begin{figure*}[t]
    \centering
    \includegraphics[width=1.9\columnwidth]{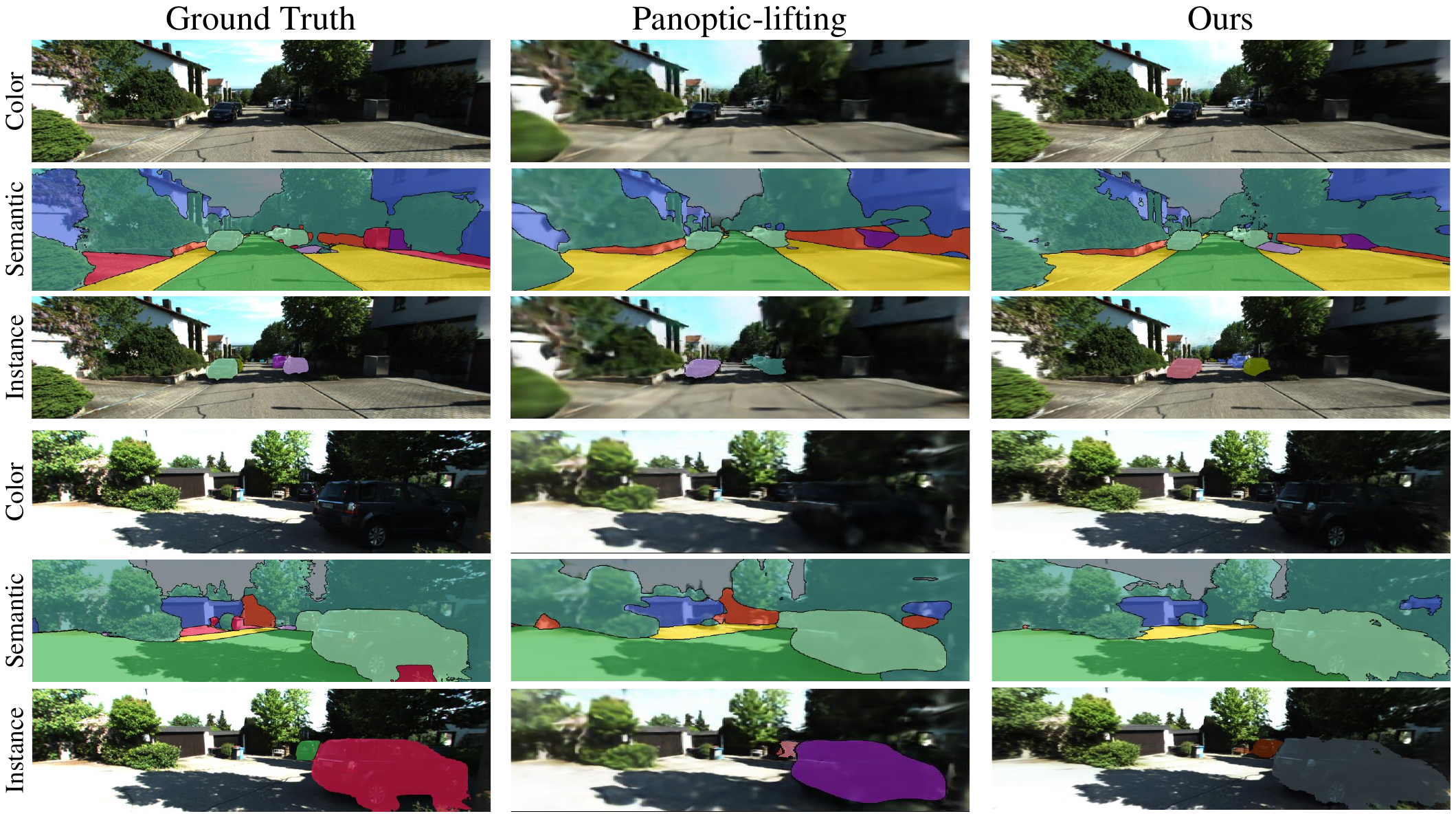}
    \caption{Illustration of Qualitative Results on the KITTI-360 Dataset. The proposed method demonstrates enhanced performance in outdoor scenes, leveraging the reparameterized domain and joint learning within the neural radiance field framework.}
    \label{fig8-vis-kitti360}
\end{figure*}

The qualitative experimental results of the proposed method are illustrated in Figures \ref{fig7-vis-replica} and \ref{fig8-vis-kitti360}. High-quality reconstruction and representation of both indoor and outdoor scenes are performed, leveraging the robust capabilities of the implicit scene representation and understanding model. Additionally, the integration of feature-prior-guided regularization and the segmentation consistency loss function enables the proposed method to achieve more accurate and 3D-consistent panoptic segmentation results.

\subsection{Ablation Study on 3D Scene Panoptic Understanding}

\input{tables/table_06_ablation.tex}

To further validate the effectiveness of each improvement, an ablation study is conducted on the ai\_001\_008 sequence of the HyperSim dataset \cite{roberts2021hypersim}. The effects of each main module within the proposed method are detailed in Table~\ref{table-ablation}. This ablation study primarily examines the impact of three main modules: the implicit scene representation and understanding model $\mathcal{S}$ (Impl.), the segmentation consistency loss function $\mathcal{L}_\text{seg}$ (Seg.), and the feature-prior-guided regularization $\mathcal{L}_\text{feat}$ (Reg.). Four variants are considered:
\begin{enumerate}
\item The baseline variant (Base), which employs an MLP to construct the implicit scene representation and understanding model;
\item The variant (B.$\mathcal{S}$), which incorporates the implicit scene representation and understanding model $\mathcal{S}$ into the baseline variant;
\item The variant (B.$\mathcal{S.L}$), which adds the segmentation consistency loss function $\mathcal{L}_\text{seg}$ to the B.$\mathcal{S}$ variant;
\item The variant (B.$\mathcal{S.R}$), which integrates the feature-prior-guided regularization $\mathcal{L}_\text{feat}$ into the B.$\mathcal{S}$ variant.
\end{enumerate}

The results of the ablation study are presented in Table~\ref{table-ablation}. The scores indicate that the main modules introduced in the proposed method significantly enhance performance across different combinations. Firstly, the B.$\mathcal{S}$ variant demonstrates a substantial improvement in appearance geometry representation compared to the Base, as well as enhancements in semantic and panoptic segmentation metrics. This suggests that the implicit scene representation and understanding model $\mathcal{S}$ enhances the model’s ability to represent scenes and provides a solid foundation for 3D semantic and instance segmentation. Secondly, by comparing the results of B.$\mathcal{S.L}$ with B.$\mathcal{S}$, it is evident that the segmentation consistency loss function $\mathcal{L}_\text{seg}$ further improves the 3D consistency and accuracy of semantic and instance segmentation. Additionally, B.$\mathcal{S.R}$ shows significant improvement over B.$\mathcal{S}$ due to the introduction of feature-prior-guided regularization, which enhances the correlation between appearance geometry representation and semantic instance representation, thereby facilitating coherent representation and understanding of the target scene. Finally, when all main modules are employed, the proposed method achieves effective representation of the target scene in terms of appearance, geometry, semantics, and instances.


\section{Conclusion}
This paper presents a novel perceptual-prior-guided 3D scene representation and panoptic understanding method, which achieves the reconstruction and panoptic understanding of indoor and outdoor target scenes. The accuracy challenges in 3D mapping during the construction of implicit panoptic maps, the complex characteristics of target scenes, and the noise in panoptic pseudo-labels are addressed by the joint optimization of geometry, appearance, semantics, and instances. The proposed method utilizes 2D image panoptic segmentation and employs a scene neural radiance field panoptic understanding implicit model to simultaneously reconstruct and understand the target scene. Perceptual features from the 2D image panoptic understanding model are introduced as prior information guidance during the learning process, synchronizing the appearance and geometric learning with the scene panoptic understanding process. This approach results in scene representation and panoptic understanding with higher accuracy and 3D consistency. The proposed method demonstrates excellent reconstruction and segmentation results in scene representation and panoptic understanding through multiple benchmarks against advanced baseline methods.

\bibliographystyle{IEEEtran}
\bibliography{IEEEabrv,references}

\begin{IEEEbiography}[{\includegraphics[width=1in,height=1.25in,clip,keepaspectratio]{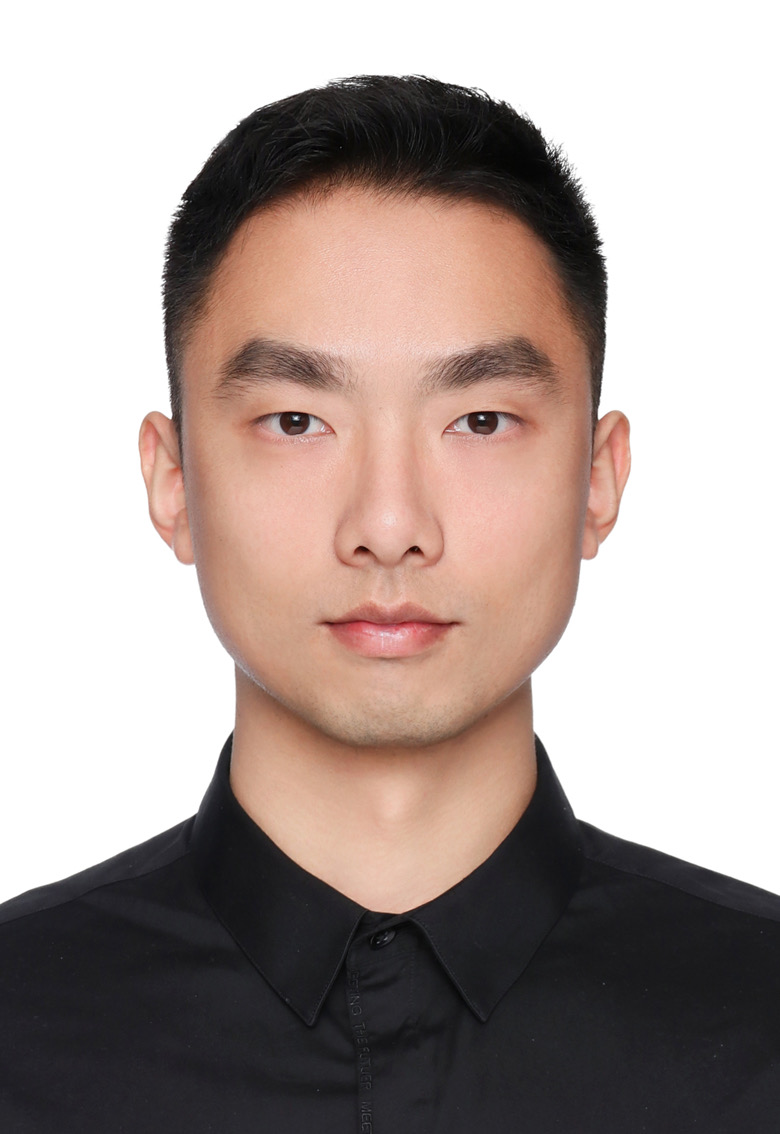}}]{Shenghao Li}
received the B.S. degree in Mechanical Design, Manufacture, and Automation from East China University of Science and Technology in 2017 and the M.S. degree in Mechanical Engineering in 2020 from East China University of Science and Technology, Shanghai. He is pursuing a Ph.D. degree in Control Science and Engineering at Shanghai Jiao Tong University, Shanghai, China. His current research interests include local feature learning, visual localization, and visual-inertial SLAM.
\end{IEEEbiography}

\end{document}

%% file: tables/table_02_replica.tex
\begin{table}[t]
	\caption{Evaluation Results on Replica}
	\label{table-benchmark-replica}
	\begin{center}
		\begin{threeparttable}
			\setlength{\tabcolsep}{1.2mm}
			\begin{tabular}{l c c c c c}
				\toprule
				\multirow{2}{*}{Methods}			& \multicolumn{5}{c}{Average Metrics}\\
				\cmidrule(lr){2-6}
				&mIOU$\uparrow$ &PQ$^\text{scene}$$\uparrow$ &SQ$^\text{scene}$$\uparrow$ &RQ$^\text{scene}$$\uparrow$ &PSNR$\uparrow$\\
				\midrule
				Mask2Former\cite{cheng2022masked}$\dagger$		                &52.4&-&-&-&-\\
				SemanticNeRF\cite{zhi2021place}$\dagger$ 	                    &58.5&-&-&-&24.8\\
				DM-NeRF\cite{wang2022dm}$\dagger$ 	                            &56.0&44.1&58.7&47.7&26.9\\
				PNF\cite{kundu2022panoptic}$\dagger$ 	                        &51.5&41.1&53.6&44.1&29.8\\
				PNF+GT.Box\cite{kundu2022panoptic}$\dagger$ 	                    &54.8&52.5&{\color{blue}\underline{62.2}}&50.8&{\color{blue}\underline{31.6}}\\
				Panoptic-Lifting\cite{siddiqui2023panoptic}$\dagger$ 	        &{\color{blue}\underline{67.2}}&{\color{blue}\underline{57.9}}&{\color{red}\textbf{69.1}}&{\color{blue}\underline{63.6}}&29.6\\
				\midrule
				Ours								                    &{\color{red}\textbf{67.8}}&{\color{red}\textbf{58.1}}&61.6&{\color{red}\textbf{67.6}}&{\color{red}\textbf{36.6}}\\
				\bottomrule
			\end{tabular}
			\begin{tablenotes}
				\item[*] The best and the second best results are marked as {\color{red}\textbf{red}} and {\color{blue}\underline{blue}}, respectively. $\dagger$ marks the results from~\cite{siddiqui2023panoptic}.
			\end{tablenotes}
		\end{threeparttable}
	\end{center}
\end{table}

%% file: tables/table_01_hypersim.tex
\begin{table}[t]
	\caption{Evaluation Results on HyperSim}
	\label{table-benchmark-hypersim}
	\begin{center}
		\begin{threeparttable}
			\setlength{\tabcolsep}{1.2mm}
			\begin{tabular}{l c c c c c}
				\toprule
				\multirow{2}{*}{Methods}			& \multicolumn{5}{c}{Average Metrics}\\
				\cmidrule(lr){2-6}
				&mIOU$\uparrow$ &PQ$^\text{scene}$$\uparrow$ &SQ$^\text{scene}$$\uparrow$ &RQ$^\text{scene}$$\uparrow$ &PSNR$\uparrow$\\
				\midrule
				Mask2Former\cite{cheng2022masked}$\dagger$		                &53.9&-&-&-&-\\
				SemanticNeRF\cite{zhi2021place}$\dagger$ 	                    &58.9&-&-&-&26.6\\
				DM-NeRF\cite{wang2022dm}$\dagger$ 	                            &57.6&51.6&62.1&55.5&28.1\\
				PNF\cite{kundu2022panoptic}$\dagger$ 	                        &50.3&44.8&55.3&47.5&27.4\\
				PNF+GT.Box\cite{kundu2022panoptic}$\dagger$ 	                    &58.7&47.6&68.2&53.4&28.1\\
				Panoptic-Lifting\cite{siddiqui2023panoptic}$\dagger$ 	        &{\color{red}\textbf{67.8}}&{\color{blue}\underline{60.1}}&{\color{blue}\underline{70.4}}&{\color{blue}\underline{64.3}}&{\color{blue}\underline{30.1}}\\
				\midrule
				Ours								                    &{\color{blue}\underline{66.3}}&{\color{red}\textbf{67.2}}&{\color{red}\textbf{72.5}}&{\color{red}\textbf{79.5}}&{\color{red}\textbf{31.5}}\\
				\bottomrule
			\end{tabular}
			\begin{tablenotes}
				\item[*] The best and the second best results are marked as {\color{red}\textbf{red}} and {\color{blue}\underline{blue}}, respectively. $\dagger$ marks the results from~\cite{siddiqui2023panoptic}.
			\end{tablenotes}
		\end{threeparttable}
	\end{center}
\end{table}

%% file: tables/table_03_scannet.tex
\begin{table}[t]
	\caption{Evaluation Results on ScanNet}
	\label{table-benchmark-scannet}
	\begin{center}
		\begin{threeparttable}
			\setlength{\tabcolsep}{1.2mm}
			\begin{tabular}{l c c c c c}
				\toprule
				\multirow{2}{*}{Methods}			& \multicolumn{5}{c}{Average Metrics}\\
				\cmidrule(lr){2-6}
				&mIOU$\uparrow$ &PQ$^\text{scene}$$\uparrow$ &SQ$^\text{scene}$$\uparrow$ &RQ$^\text{scene}$$\uparrow$ &PSNR$\uparrow$\\
				\midrule
				Mask2Former\cite{cheng2022masked}$\dagger$		                &46.7&-&-&-&-\\
				SemanticNeRF\cite{zhi2021place}$\dagger$ 	                    &59.2&-&-&-&26.6\\
				DM-NeRF\cite{wang2022dm}$\dagger$ 	                            &49.5&41.7&53.3&46.1&27.5\\
				PNF\cite{kundu2022panoptic}$\dagger$ 	                        &53.9&48.3&63.0&50.7&26.7\\
				PNF+GT.Box\cite{kundu2022panoptic}$\dagger$ 	                    &58.7&54.3&70.0&55.9&26.8\\
				Panoptic-Lifting\cite{siddiqui2023panoptic}$\dagger$ 	        &{\color{blue}\underline{65.2}}&{\color{red}\textbf{58.9}}&{\color{red}\textbf{73.5}}&{\color{blue}\underline{65.0}}&{\color{blue}\underline{28.5}}\\
				\midrule
				Ours								                    &{\color{red}\textbf{66.2}}&{\color{blue}\underline{57.0}}&{\color{blue}\underline{72.9}}&{\color{red}\textbf{67.4}}&{\color{red}\textbf{28.9}}\\
				\bottomrule
			\end{tabular}
			\begin{tablenotes}
				\item[*] The best and the second best results are marked as {\color{red}\textbf{red}} and {\color{blue}\underline{blue}}, respectively. $\dagger$ marks the results from~\cite{siddiqui2023panoptic}.
			\end{tablenotes}
		\end{threeparttable}
	\end{center}
\end{table}

%% file: tables/table_05_kitti360_cus.tex
\begin{table}[t]
	\caption{Evaluation Results on KITTI-360}
	\label{table-benchmark-kitti360-cus}
	\begin{center}
		\begin{threeparttable}
			\setlength{\tabcolsep}{1.2mm}
			\begin{tabular}{l c c c c c}
				\toprule
				\multirow{2}{*}{Methods}			& \multicolumn{5}{c}{Average Metrics}\\
				\cmidrule(lr){2-6}
				&mIOU$\uparrow$ &PQ$^\text{scene}$$\uparrow$ &SQ$^\text{scene}$$\uparrow$ &RQ$^\text{scene}$$\uparrow$ &PSNR$\uparrow$\\
				\midrule
				Mask2Former\cite{cheng2022masked}		                &55.4&-&-&-&-\\
				NeRF\cite{mildenhall2021nerf}		                    &46.4&40.7&42.9&58.4&17.0\\
				MipNeRF\cite{barron2021mipnerf}		                    &47.5&41.8&44.0&58.5&17.9\\
				Panoptic-Lifting\cite{siddiqui2023panoptic} 	        &{\color{blue}\underline{50.5}}&{\color{blue}\underline{45.6}}&{\color{red}\textbf{61.8}}&{\color{blue}\underline{47.4}}&{\color{blue}\underline{20.1}}\\
				\midrule
				Ours								                    &{\color{red}\textbf{63.4}}&{\color{red}\textbf{57.7}}&{\color{blue}\underline{60.0}}&{\color{red}\textbf{70.8}}&{\color{red}\textbf{21.6}}\\
				\bottomrule
			\end{tabular}
			\begin{tablenotes}
				\item[*] The best and the second best results are marked as {\color{red}\textbf{red}} and {\color{blue}\underline{blue}}, respectively.
			\end{tablenotes}
		\end{threeparttable}
	\end{center}
\end{table}

%% file: tables/table_07_efficiency.tex
\begin{table}[t]
	\caption{Runtime Analysis of 3D Scene Representation and Panoptic Understanding}
	\label{table-runtime-analysis}
	\begin{center}
		\begin{threeparttable}
			\setlength{\tabcolsep}{10mm}
			\begin{tabular}{l c}
				\toprule
				Methods			& Runtime[ms]\\
				\midrule
				NeRF\cite{mildenhall2021nerf} 	            &63.7\\
				MipNeRF\cite{barron2021mipnerf} 	        &74.9\\
				SemantiNeRF\cite{siddiqui2023panoptic} 	    &94.5\\
				DM-NeRF\cite{siddiqui2023panoptic} 	        &96.4\\
				Panoptic-Lifting\cite{siddiqui2023panoptic} &{\color{blue}\underline{28.5}}\\
				\midrule
				Ours					                    &{\color{red}\textbf{26.7}}\\
				\bottomrule
			\end{tabular}
            \begin{tablenotes}
				\item[*] The best and the second best results are marked as {\color{red}\textbf{red}} and {\color{blue}\underline{blue}}, respectively. Runtime is computed by rendering 2048 rays.
			\end{tablenotes}
		\end{threeparttable}
	\end{center}
\end{table}

%% file: tables/table_06_ablation.tex
\begin{table*}[t]
	\caption{Ablation Study Results on HyperSim}
	\label{table-ablation}
	\begin{center}
		\begin{threeparttable}
			\setlength{\tabcolsep}{5mm}
			\begin{tabular}{l c c c c c c c c}
				\toprule
				\multirow{2}{*}{Methods}		&\multicolumn{3}{c}{Modules}	&\multicolumn{5}{c}{Average Metrics} \\
				\cmidrule(lr){2-4}\cmidrule(lr){5-9}
				&Impl. &Seg. &Reg.	&mIOU$\uparrow$ &PQ$^\text{scene}$$\uparrow$ &SQ$^\text{scene}$$\uparrow$ &RQ$^\text{scene}$$\uparrow$ &PSNR$\uparrow$ \\
				\midrule
				Base	& -				&-					&-					                                    &58.7&55.4&60.2&61.4&29.5\\
				B.$\mathcal{S}$	& $\checkmark$				&-					&-					                &60.6&65.6&65.6&75.0&32.0\\
				B.$\mathcal{S.L}$	& $\checkmark$	& $\checkmark$		&-					                        &{\color{blue}\underline{64.6}}&{\color{blue}\underline{76.7}}&{\color{blue}\underline{78.2}}&{\color{blue}\underline{83.6}}&{\color{red}\textbf{32.8}}\\
				B.$\mathcal{S.R}$	& $\checkmark$	& - & $\checkmark$					                        &62.0&71.3&75.4&82.1&32.4\\
				\midrule
				Ours	& $\checkmark$	& $\checkmark$			&$\checkmark$   &{\color{red}\textbf{65.2}}&{\color{red}\textbf{79.2}}&{\color{red}\textbf{80.8}}&{\color{red}\textbf{84.1}}&{\color{blue}\underline{32.7}}\\	
				\bottomrule
			\end{tabular}
			\begin{tablenotes}
				\item[*]  The best and the second best results are marked as {\color{red}\textbf{red}} and {\color{blue}\underline{blue}}, respectively. The ablation study is performed on the ai\_001\_008 sequence of HyperSim.
			\end{tablenotes}
		\end{threeparttable}
	\end{center}
\end{table*}

%% file: manuscript.bbl
\begin{thebibliography}{10}
\providecommand{\url}[1]{#1}
\csname url@samestyle\endcsname
\providecommand{\newblock}{\relax}
\providecommand{\bibinfo}[2]{#2}
\providecommand{\BIBentrySTDinterwordspacing}{\spaceskip=0pt\relax}
\providecommand{\BIBentryALTinterwordstretchfactor}{4}
\providecommand{\BIBentryALTinterwordspacing}{\spaceskip=\fontdimen2\font plus
\BIBentryALTinterwordstretchfactor\fontdimen3\font minus \fontdimen4\font\relax}
\providecommand{\BIBforeignlanguage}[2]{{%
\expandafter\ifx\csname l@#1\endcsname\relax
\typeout{** WARNING: IEEEtran.bst: No hyphenation pattern has been}%
\typeout{** loaded for the language `#1'. Using the pattern for}%
\typeout{** the default language instead.}%
\else
\language=\csname l@#1\endcsname
\fi
#2}}
\providecommand{\BIBdecl}{\relax}
\BIBdecl

\bibitem{cheng2022masked}
B.~Cheng, I.~Misra, A.~G. Schwing, A.~Kirillov, and R.~Girdhar, ``Masked-attention mask transformer for universal image segmentation,'' in \emph{2022 IEEE/CVF Conference on Computer Vision and Pattern Recognition (CVPR)}, 2022, pp. 1290--1299.

\bibitem{li2023mask}
F.~Li, H.~Zhang, H.~Xu, S.~Liu, L.~Zhang, L.~M. Ni, and H.-Y. Shum, ``Mask dino: Towards a unified transformer-based framework for object detection and segmentation,'' in \emph{2023 IEEE/CVF Conference on Computer Vision and Pattern Recognition (CVPR)}, 2023, pp. 3041--3050.

\bibitem{yu2022k}
Q.~Yu, H.~Wang, S.~Qiao, M.~Collins, Y.~Zhu, H.~Adam, A.~Yuille, and L.-C. Chen, ``k-means mask transformer,'' in \emph{2022 European Conference on Computer Vision (ECCV)}.\hskip 1em plus 0.5em minus 0.4em\relax Springer, 2022, pp. 288--307.

\bibitem{kirillov2019panoptic}
A.~Kirillov, K.~He, R.~Girshick, C.~Rother, and P.~Doll{\'a}r, ``Panoptic segmentation,'' in \emph{2019 IEEE/CVF Conference on Computer Vision and Pattern Recognition (CVPR)}, 2019, pp. 9404--9413.

\bibitem{fu2022panoptic}
X.~Fu, S.~Zhang, T.~Chen, Y.~Lu, L.~Zhu, X.~Zhou, A.~Geiger, and Y.~Liao, ``Panoptic nerf: 3d-to-2d label transfer for panoptic urban scene segmentation,'' in \emph{2022 International Conference on 3D Vision (3DV)}, 2022, pp. 301--111.

\bibitem{kundu2022panoptic}
A.~Kundu, K.~Genova, X.~Yin, A.~Fathi, C.~Pantofaru, L.~J. Guibas, A.~Tagliasacchi, F.~Dellaert, and T.~Funkhouser, ``Panoptic neural fields: A semantic object-aware neural scene representation,'' in \emph{2022 IEEE/CVF Conference on Computer Vision and Pattern Recognition (CVPR)}, 2022, pp. 12\,871--12\,881.

\bibitem{wang2022dm}
B.~Wang, L.~Chen, and B.~Yang, ``Dm-nerf: 3d scene geometry decomposition and manipulation from 2d images,'' \emph{arXiv preprint arXiv:2208.07227}, 2022.

\bibitem{zhi2021place}
S.~Zhi, T.~Laidlow, S.~Leutenegger, and A.~J. Davison, ``In-place scene labelling and understanding with implicit scene representation,'' in \emph{2021 IEEE/CVF International Conference on Computer Vision (ICCV)}, 2021, pp. 15\,838--15\,847.

\bibitem{siddiqui2023panoptic}
Y.~Siddiqui, L.~Porzi, S.~R. Bul{\`o}, N.~M{\"u}ller, M.~Nie{\ss}ner, A.~Dai, and P.~Kontschieder, ``Panoptic lifting for 3d scene understanding with neural fields,'' in \emph{2023 IEEE/CVF Conference on Computer Vision and Pattern Recognition (CVPR)}, 2023, pp. 9043--9052.

\bibitem{barron2021mipnerf}
J.~T. Barron, B.~Mildenhall, M.~Tancik, P.~Hedman, R.~Martin-Brualla, and P.~P. Srinivasan, ``{Mip-NeRF}: A multiscale representation for anti-aliasing neural radiance fields,'' in \emph{2021 International Conference on Computer Vision (ICCV)}, 2021, pp. 5835--5844.

\bibitem{barron2022mipnerf360}
J.~T. Barron, B.~Mildenhall, D.~Verbin, P.~P. Srinivasan, and P.~Hedman, ``{Mip-NeRF 360}: Unbounded anti-aliased neural radiance fields,'' in \emph{2022 IEEE/CVF Conference on Computer Vision and Pattern Recognition (CVPR)}, 2022, pp. 5460--5469.

\bibitem{caesar2018coco}
H.~Caesar, J.~Uijlings, and V.~Ferrari, ``Coco-stuff: Thing and stuff classes in context,'' in \emph{2018 IEEE/CVF Conference on Computer Vision and Pattern Recognition (CVPR)}, 2018, pp. 1209--1218.

\bibitem{hermans2014dense}
A.~Hermans, G.~Floros, and B.~Leibe, ``Dense 3d semantic mapping of indoor scenes from rgb-d images,'' in \emph{2014 IEEE International Conference on Robotics and Automation (ICRA)}.\hskip 1em plus 0.5em minus 0.4em\relax IEEE, 2014, pp. 2631--2638.

\bibitem{mccormac2017semanticfusion}
J.~McCormac, A.~Handa, A.~Davison, and S.~Leutenegger, ``Semanticfusion: Dense 3d semantic mapping with convolutional neural networks,'' in \emph{2017 IEEE International Conference on Robotics and automation (ICRA)}.\hskip 1em plus 0.5em minus 0.4em\relax IEEE, 2017, pp. 4628--4635.

\bibitem{runz2018maskfusion}
M.~Runz, M.~Buffier, and L.~Agapito, ``Maskfusion: Real-time recognition, tracking and reconstruction of multiple moving objects,'' in \emph{2018 IEEE International Symposium on Mixed and Augmented Reality (ISMAR)}, 2018, pp. 10--20.

\bibitem{mccormac2018fusion++}
J.~McCormac, R.~Clark, M.~Bloesch, A.~Davison, and S.~Leutenegger, ``Fusion++: Volumetric object-level slam,'' in \emph{2018 international conference on 3D vision (3DV)}.\hskip 1em plus 0.5em minus 0.4em\relax IEEE, 2018, pp. 32--41.

\bibitem{he2017mask}
K.~He, G.~Gkioxari, P.~Doll{\'a}r, and R.~Girshick, ``Mask r-cnn,'' in \emph{2017 IEEE International Conference on Computer Vision (CVPR)}, 2017, pp. 2961--2969.

\bibitem{narita2019panopticfusion}
G.~Narita, T.~Seno, T.~Ishikawa, and Y.~Kaji, ``Panopticfusion: Online volumetric semantic mapping at the level of stuff and things,'' in \emph{2019 IEEE/RSJ International Conference on Intelligent Robots and Systems (IROS)}.\hskip 1em plus 0.5em minus 0.4em\relax IEEE, 2019, pp. 4205--4212.

\bibitem{han2020occuseg}
L.~Han, T.~Zheng, L.~Xu, and L.~Fang, ``Occuseg: Occupancy-aware 3d instance segmentation,'' in \emph{2020 IEEE/CVF Conference on Computer Vision and Pattern Recognition (CVPR)}, 2020, pp. 2940--2949.

\bibitem{qi2017pointnet++}
C.~R. Qi, L.~Yi, H.~Su, and L.~J. Guibas, ``Pointnet++: Deep hierarchical feature learning on point sets in a metric space,'' \emph{Advances in Neural Information Processing Systems}, vol.~30, pp. 5100--5109, 2017.

\bibitem{huang2019texturenet}
J.~Huang, H.~Zhang, L.~Yi, T.~Funkhouser, M.~Nie{\ss}ner, and L.~J. Guibas, ``Texturenet: Consistent local parametrizations for learning from high-resolution signals on meshes,'' in \emph{2019 IEEE/CVF Conference on Computer Vision and Pattern Recognition (CVPR)}, 2019, pp. 4440--4449.

\bibitem{schult2020dualconvmesh}
J.~Schult, F.~Engelmann, T.~Kontogianni, and B.~Leibe, ``Dualconvmesh-net: Joint geodesic and euclidean convolutions on 3d meshes,'' in \emph{2020 IEEE/CVF Conference on Computer Vision and Pattern Recognition (CVPR)}, 2020, pp. 8612--8622.

\bibitem{hu2021vmnet}
Z.~Hu, X.~Bai, J.~Shang, R.~Zhang, J.~Dong, X.~Wang, G.~Sun, H.~Fu, and C.-L. Tai, ``Vmnet: Voxel-mesh network for geodesic-aware 3d semantic segmentation,'' in \emph{2021 IEEE/CVF International Conference on Computer Vision (CVPR)}, 2021, pp. 15\,488--15\,498.

\bibitem{hu2021bidirectional}
W.~Hu, H.~Zhao, L.~Jiang, J.~Jia, and T.-T. Wong, ``Bidirectional projection network for cross dimension scene understanding,'' in \emph{2021 IEEE/CVF Conference on Computer Vision and Pattern Recognition (CVPR)}, 2021, pp. 14\,373--14\,382.

\bibitem{ronneberger2015unet}
O.~Ronneberger, P.~Fischer, and T.~Brox, ``U-net: Convolutional networks for biomedical image segmentation,'' in \emph{2015 International Conference on Medical Image Computing and Computer-Assisted Intervention}.\hskip 1em plus 0.5em minus 0.4em\relax Springer, 2015, pp. 234--241.

\bibitem{choy20194d}
C.~Choy, J.~Gwak, and S.~Savarese, ``4d spatio-temporal convnets: Minkowski convolutional neural networks,'' in \emph{2019 IEEE/CVF Conference on Computer Vision and Pattern Recognition (CVPR)}, 2019, pp. 3075--3084.

\bibitem{nekrasov2021mix3d}
A.~Nekrasov, J.~Schult, O.~Litany, B.~Leibe, and F.~Engelmann, ``Mix3d: Out-of-context data augmentation for 3d scenes,'' in \emph{2021 International Conference on 3D Vision (3DV)}.\hskip 1em plus 0.5em minus 0.4em\relax IEEE, 2021, pp. 116--125.

\bibitem{radford2021learning}
A.~Radford, J.~W. Kim, C.~Hallacy, A.~Ramesh, G.~Goh, S.~Agarwal, G.~Sastry, A.~Askell, P.~Mishkin, J.~Clark \emph{et~al.}, ``Learning transferable visual models from natural language supervision,'' in \emph{2021 International Conference on Machine Learning (ICML)}.\hskip 1em plus 0.5em minus 0.4em\relax PMLR, 2021, pp. 8748--8763.

\bibitem{li2022panoptic}
J.~Li, X.~He, Y.~Wen, Y.~Gao, X.~Cheng, and D.~Zhang, ``Panoptic-phnet: Towards real-time and high-precision lidar panoptic segmentation via clustering pseudo heatmap,'' in \emph{2022 IEEE/CVF Conference on Computer Vision and Pattern Recognition (CVPR)}, 2022, pp. 11\,809--11\,818.

\bibitem{robert2022learning}
D.~Robert, B.~Vallet, and L.~Landrieu, ``Learning multi-view aggregation in the wild for large-scale 3d semantic segmentation,'' in \emph{2022 IEEE/CVF Conference on Computer Vision and Pattern Recognition (CVPR)}, 2022, pp. 5575--5584.

\bibitem{wu20153d}
Z.~Wu, S.~Song, A.~Khosla, F.~Yu, L.~Zhang, X.~Tang, and J.~Xiao, ``3d shapenets: A deep representation for volumetric shapes,'' in \emph{2015 IEEE/CVF Conference on Computer Vision and Pattern Recognition (CVPR)}, 2015, pp. 1912--1920.

\bibitem{chen2020scanrefer}
D.~Z. Chen, A.~X. Chang, and M.~Nie{\ss}ner, ``Scanrefer: 3d object localization in rgb-d scans using natural language,'' in \emph{2020 European Conference on Computer Vision (ECCV)}.\hskip 1em plus 0.5em minus 0.4em\relax Springer, 2020, pp. 202--221.

\bibitem{niemeyer2021giraffe}
M.~Niemeyer and A.~Geiger, ``{GIRAFFE}: Representing scenes as compositional generative neural feature fields,'' in \emph{2021 IEEE/CVF Conference on Computer Vision and Pattern Recognition (CVPR)}, 2021, pp. 11\,448--11\,459.

\bibitem{sun2020scalability}
P.~Sun, H.~Kretzschmar, X.~Dotiwalla, A.~Chouard, V.~Patnaik, P.~Tsui, J.~Guo, Y.~Zhou, Y.~Chai, B.~Caine \emph{et~al.}, ``Scalability in perception for autonomous driving: Waymo open dataset,'' in \emph{2020 IEEE/CVF Conference on Computer Vision and Pattern Recognition (CVPR)}, 2020, pp. 2446--2454.

\bibitem{armeni2017joint}
I.~Armeni, S.~Sax, A.~R. Zamir, and S.~Savarese, ``Joint 2d-3d-semantic data for indoor scene understanding,'' \emph{arXiv preprint arXiv:1702.01105}, 2017.

\bibitem{behley2019semantickitti}
J.~Behley, M.~Garbade, A.~Milioto, J.~Quenzel, S.~Behnke, C.~Stachniss, and J.~Gall, ``Semantickitti: A dataset for semantic scene understanding of lidar sequences,'' in \emph{2019 IEEE/CVF International Conference on Computer Vision (CVPR)}, 2019, pp. 9297--9307.

\bibitem{liao2023kitti}
Y.~Liao, J.~Xie, and A.~Geiger, ``Kitti-360: A novel dataset and benchmarks for urban scene understanding in 2d and 3d,'' \emph{IEEE Transactions on Pattern Analysis and Machine Intelligence}, vol.~45, no.~3, pp. 3292--3310, 2023.

\bibitem{deng20213d}
S.~Deng, X.~Xu, C.~Wu, K.~Chen, and K.~Jia, ``3d affordancenet: A benchmark for visual object affordance understanding,'' in \emph{2021 IEEE/CVF Conference on Computer Vision and Pattern Recognition (CVPR)}, 2021, pp. 1778--1787.

\bibitem{li2019putting}
X.~Li, S.~Liu, K.~Kim, X.~Wang, M.-H. Yang, and J.~Kautz, ``Putting humans in a scene: Learning affordance in 3d indoor environments,'' in \emph{2019 IEEE/CVF Conference on Computer Vision and Pattern Recognition (CVPR)}, 2019, pp. 12\,368--12\,376.

\bibitem{wang2021synthesizing}
J.~Wang, H.~Xu, J.~Xu, S.~Liu, and X.~Wang, ``Synthesizing long-term 3d human motion and interaction in 3d scenes,'' in \emph{2021 IEEE/CVF Conference on Computer Vision and Pattern Recognition (CVPR)}, 2021, pp. 9401--9411.

\bibitem{genova2021learning}
K.~Genova, X.~Yin, A.~Kundu, C.~Pantofaru, F.~Cole, A.~Sud, B.~Brewington, B.~Shucker, and T.~Funkhouser, ``Learning 3d semantic segmentation with only 2d image supervision,'' in \emph{2021 International Conference on 3D Vision (3DV)}.\hskip 1em plus 0.5em minus 0.4em\relax IEEE, 2021, pp. 361--372.

\bibitem{kundu2020virtual}
A.~Kundu, X.~Yin, A.~Fathi, D.~Ross, B.~Brewington, T.~Funkhouser, and C.~Pantofaru, ``Virtual multi-view fusion for 3d semantic segmentation,'' in \emph{2020 European Conference on Computer Vision (ECCV)}.\hskip 1em plus 0.5em minus 0.4em\relax Springer, 2020, pp. 518--535.

\bibitem{sautier2022image}
C.~Sautier, G.~Puy, S.~Gidaris, A.~Boulch, A.~Bursuc, and R.~Marlet, ``Image-to-lidar self-supervised distillation for autonomous driving data,'' in \emph{2022 IEEE/CVF Conference on Computer Vision and Pattern Recognition (CVPR)}, 2022, pp. 9891--9901.

\bibitem{wang2022detr3d}
Y.~Wang, V.~C. Guizilini, T.~Zhang, Y.~Wang, H.~Zhao, and J.~Solomon, ``Detr3d: 3d object detection from multi-view images via 3d-to-2d queries,'' in \emph{Conference on Robot Learning}.\hskip 1em plus 0.5em minus 0.4em\relax PMLR, 2022, pp. 180--191.

\bibitem{liu2021contrastive}
Y.~Liu, Q.~Fan, S.~Zhang, H.~Dong, T.~Funkhouser, and L.~Yi, ``Contrastive multimodal fusion with tupleinfonce,'' in \emph{2021 IEEE/CVF International Conference on Computer Vision (CVPR)}, 2021, pp. 754--763.

\bibitem{jia2021scaling}
C.~Jia, Y.~Yang, Y.~Xia, Y.-T. Chen, Z.~Parekh, H.~Pham, Q.~Le, Y.-H. Sung, Z.~Li, and T.~Duerig, ``Scaling up visual and vision-language representation learning with noisy text supervision,'' in \emph{2021 International Conference on Machine Learning (ICML)}.\hskip 1em plus 0.5em minus 0.4em\relax PMLR, 2021, pp. 4904--4916.

\bibitem{alayrac2022flamingo}
J.-B. Alayrac, J.~Donahue, P.~Luc, A.~Miech, I.~Barr, Y.~Hasson, K.~Lenc, A.~Mensch, K.~Millican, M.~Reynolds \emph{et~al.}, ``Flamingo: a visual language model for few-shot learning,'' \emph{Advances in Neural Information Processing Systems}, vol.~35, pp. 23\,716--23\,736, 2022.

\bibitem{gu2021open}
X.~Gu, T.-Y. Lin, W.~Kuo, and Y.~Cui, ``Open-vocabulary object detection via vision and language knowledge distillation,'' in \emph{2022 International Conference on Learning Representations (ICLR)}, 2022.

\bibitem{ghiasi2022scaling}
G.~Ghiasi, X.~Gu, Y.~Cui, and T.-Y. Lin, ``Scaling open-vocabulary image segmentation with image-level labels,'' in \emph{2022 European Conference on Computer Vision (ECCV)}.\hskip 1em plus 0.5em minus 0.4em\relax Springer, 2022, pp. 540--557.

\bibitem{he2023open}
S.~He, T.~Guo, T.~Dai, R.~Qiao, X.~Shu, B.~Ren, and S.-T. Xia, ``Open-vocabulary multi-label classification via multi-modal knowledge transfer,'' in \emph{Proceedings of the AAAI Conference on Artificial Intelligence}, vol.~37, no.~1, 2023, pp. 808--816.

\bibitem{ma2022open}
C.~Ma, Y.~Yang, Y.~Wang, Y.~Zhang, and W.~Xie, ``Open-vocabulary semantic segmentation with frozen vision-language models,'' \emph{arXiv preprint arXiv:2210.15138}, 2022.

\bibitem{zhou2022extract}
C.~Zhou, C.~C. Loy, and B.~Dai, ``Extract free dense labels from clip,'' in \emph{2022 European Conference on Computer Vision (ECCV)}.\hskip 1em plus 0.5em minus 0.4em\relax Springer, 2022, pp. 696--712.

\bibitem{luddecke2022image}
T.~L{\"u}ddecke and A.~Ecker, ``Image segmentation using text and image prompts,'' in \emph{2022 IEEE/CVF Conference on Computer Vision and Pattern Recognition (CVPR)}, 2022, pp. 7086--7096.

\bibitem{ha2022semantic}
H.~Ha and S.~Song, ``Semantic abstraction: Open-world 3d scene understanding from 2d vision-language models,'' in \emph{Conference on Robot Learning}, 2022.

\bibitem{Peng2023OpenScene}
S.~Peng, K.~Genova, C.~M. Jiang, A.~Tagliasacchi, M.~Pollefeys, and T.~Funkhouser, ``Openscene: 3d scene understanding with open vocabularies,'' in \emph{2023 IEEE/CVF Conference on Computer Vision and Pattern Recognition (CVPR)}, 2023.

\bibitem{li2024representing}
S.~Li, Z.~Xia, and Q.~Zhao, ``Representing boundary-ambiguous scene online with scale-encoded cascaded grids and radiance field deblurring,'' \emph{IEEE Transactions on Circuits and Systems for Video Technology}, vol.~34, no.~4, pp. 2026--2040, 2024.

\bibitem{schonberger2016colmap}
J.~L. Schönberger and J.-M. Frahm, ``Structure-from-motion revisited,'' in \emph{2016 IEEE/CVF Conference on Computer Vision and Pattern Recognition (CVPR)}, 2016, pp. 4104--4113.

\bibitem{li2023s2ld}
S.~Li, Q.~Zhao, and Z.~Xia, ``Sparse-to-local-dense matching for geometry-guided correspondence estimation,'' \emph{IEEE Transactions on Image Processing}, vol.~32, pp. 3536--3551, 2023.

\bibitem{straub2019replica}
J.~Straub, T.~Whelan, L.~Ma, Y.~Chen, E.~Wijmans, S.~Green, J.~J. Engel, R.~Mur-Artal, C.~Ren, S.~Verma \emph{et~al.}, ``The replica dataset: A digital replica of indoor spaces,'' \emph{arXiv preprint arXiv:1906.05797}, 2019.

\bibitem{roberts2021hypersim}
M.~Roberts, J.~Ramapuram, A.~Ranjan, A.~Kumar, M.~A. Bautista, N.~Paczan, R.~Webb, and J.~M. Susskind, ``Hypersim: A photorealistic synthetic dataset for holistic indoor scene understanding,'' in \emph{2021 IEEE/CVF International Conference on Computer Vision (ICCV)}, 2021, pp. 10\,912--10\,922.

\bibitem{dai2017scannet}
A.~Dai, A.~X. Chang, M.~Savva, M.~Halber, T.~Funkhouser, and M.~Nie{\ss}ner, ``Scannet: Richly-annotated 3d reconstructions of indoor scenes,'' in \emph{2017 IEEE/CVF Conference on Computer Vision and Pattern Recognition (CVPR)}, 2017, pp. 5828--5839.

\bibitem{mildenhall2021nerf}
B.~Mildenhall, P.~P. Srinivasan, M.~Tancik, J.~T. Barron, R.~Ramamoorthi, and R.~Ng, ``Nerf: Representing scenes as neural radiance fields for view synthesis,'' \emph{Communications of the ACM}, vol.~65, no.~1, pp. 99--106, 2021.

\end{thebibliography}
